\documentclass[10pt,journal,compsoc]{IEEEtran}

\ifCLASSOPTIONcompsoc
  \usepackage[nocompress]{cite}
\else
  \usepackage{cite}
\fi
\ifCLASSINFOpdf
\else
\fi

\usepackage{microtype}
\usepackage{graphicx}
\usepackage{subfigure}
\usepackage{booktabs}
\usepackage{hyperref}
\usepackage{url}
\usepackage{amsmath}
\usepackage{times}
\usepackage{url}
\usepackage{latexsym}
\usepackage{graphicx}
\usepackage{pgfplots}
\usepackage{tikz}
\usepackage{amsfonts,amssymb}
\usepackage{subfigure}
\usepackage{multirow}
\usepackage{booktabs}
\usepackage{algorithm}
\usepackage{algorithmicx}  
\usepackage{algpseudocode}
\usepackage{arydshln}
\usepackage{color}
\usepackage{ulem}
\usepackage[encapsulated]{CJK}
\newcommand{\citep}{\cite}
\newcommand{\citet}{\cite}

\begin{document}

\title{To Understand Representation of Layer-aware Sequence  Encoders as  Multi-order-Graph}

\author{Sufeng Duan, Hai Zhao*
		\IEEEcompsocitemizethanks{\IEEEcompsocthanksitem{This paper was partially supported by Key Projects of National Natural Science Foundation of China (U1836222 and 61733011)(Corresponding author: Hai Zhao).}
		\IEEEcompsocthanksitem{S. Duan, H. Zhao are with the Department of Computer Science and Engineering, Shanghai Jiao Tong University, and also with Key Laboratory of Shanghai Education Commission for Intelligent Interaction and Cognitive Engineering, Shanghai Jiao Tong University. \protect\\ E-mail: 1140339019dsf@sjtu.edu.cn, zhaohai@cs.sjtu.edu.cn.}
    	}
}

\IEEEtitleabstractindextext{%
\begin{abstract}
In this paper, we propose an explanation of representation for self-attention network (SAN) based  neural sequence encoders, which regards the information captured by the model and the encoding of the model as graph structure and the generation of these graph structures respectively. The proposed explanation applies to existing works on SAN-based models and can explain the relationship among the ability to capture the structural or linguistic information, depth of model, and length of sentence, and can also be extended to other models such as recurrent neural network based models. We also propose a revisited multigraph called Multi-order-Graph (MoG) based on our explanation to model the graph structures in the SAN-based model as subgraphs in MoG and convert the encoding of SAN-based model to the generation of MoG. Based on our explanation, we further introduce a Graph-Transformer by enhancing the ability to capture multiple subgraphs of different orders and focusing on subgraphs of high orders. Experimental results on multiple neural machine translation tasks show that the Graph-Transformer can yield effective performance improvement.

\end{abstract}

}

\maketitle

\IEEEdisplaynontitleabstractindextext

\IEEEpeerreviewmaketitle

\IEEEraisesectionheading{\section{Introduction}\label{sec:introduction}}
\IEEEPARstart{T}{HE } current natural language processing (NLP) models more and more adopt an encoder-decoder framework, in which the encoder takes a sentence as input and generates the corresponding contextualized representations for the decoder for specific processing. So far, although NLP tasks with various modeling ways, generally, there are mainly three types of encoder architectures, recurrent neural network (RNN) \citep{kalchbrenner2013recurrent, DBLP:journals/corr/BahdanauCB14, DBLP:conf/nips/SutskeverVL14}, convolutional neural network (CNN), and self-attention network (SAN) from Transformer \cite{DBLP:conf/nips/VaswaniSPUJGKP17}. As a widely-used encoder architecture, the SAN facilitates all the input representations learned in a fully-connected internal structure. In this paper, we focus on the SAN-based model.

The Transformer \cite{DBLP:conf/nips/VaswaniSPUJGKP17} is the first SAN-based model proposed for neural machine translation (NMT). As the state-of-the-art NMT model, several variants of the Transformer have been proposed for further performance improvement \citep{shaw-etal-2018-self, DBLP:conf/nips/HeTXHQ0L18} and other NLP tasks such as language model \citep{DBLP:conf/naacl/DevlinCLT19}, parsing \citep{kitaev-klein-2018-constituency, zhou-zhao-2019-head}. Following works on RNN-based models with the capacity to learn structural and syntactic information  \cite{DBLP:conf/emnlp/ShiPK16,DBLP:conf/acl/BlevinsLZ18}, researchers find that SAN-based models can also extract structural or linguistic information. For example, Jawahar et al.\cite{DBLP:conf/naacl/DevlinCLT19} showed that BERT \cite{DBLP:conf/naacl/DevlinCLT19} could capture diverse information, with surface features at the bottom, syntactic features in the middle, and semantic features at the top, which means that BERT does learn some linguistics information from data. Miaschie et al.\cite{DBLP:conf/coling/MiaschiBDV20} studied the linguistic properties encoded by BERT and showed  that BERT could encode a wide range of linguistic characteristics, but it tends to lose this information when trained on specific downstream tasks. Vig and Belinkov\cite{DBLP:conf/blackboxnlp/VigB19} showed that self-attention in the language model could capture different relationships at different layers. 

These works show that SAN-based models can embed structural and linguistic information, and the information embedding ability is related to the model depth and sentence length. More detailedly, we may get intuitions as follows, (1) different layers in SAN-based models may deliver different sorts of information, (2) increasing the depth of the model can improve the performance while improvement may be tiny when the model is too deep, and (3) modeling sentence with different lengths may indicate specified but different model depths for the best performance. Naturally, we wonder \textit{Why and How SAN-based model captures various information in the type of structural or linguistic}. 
In this work, we will give an explanation and modeling method to answer the question, which can make us understand the encoding and representation from a general view. It is not more than existing related works which only focus on empirical analysis, we also make an attempt in explaining the encoding mechanism of the SAN-based model for structural or linguistic information.

\begin{figure}
    \centering
    \includegraphics[scale=0.23]{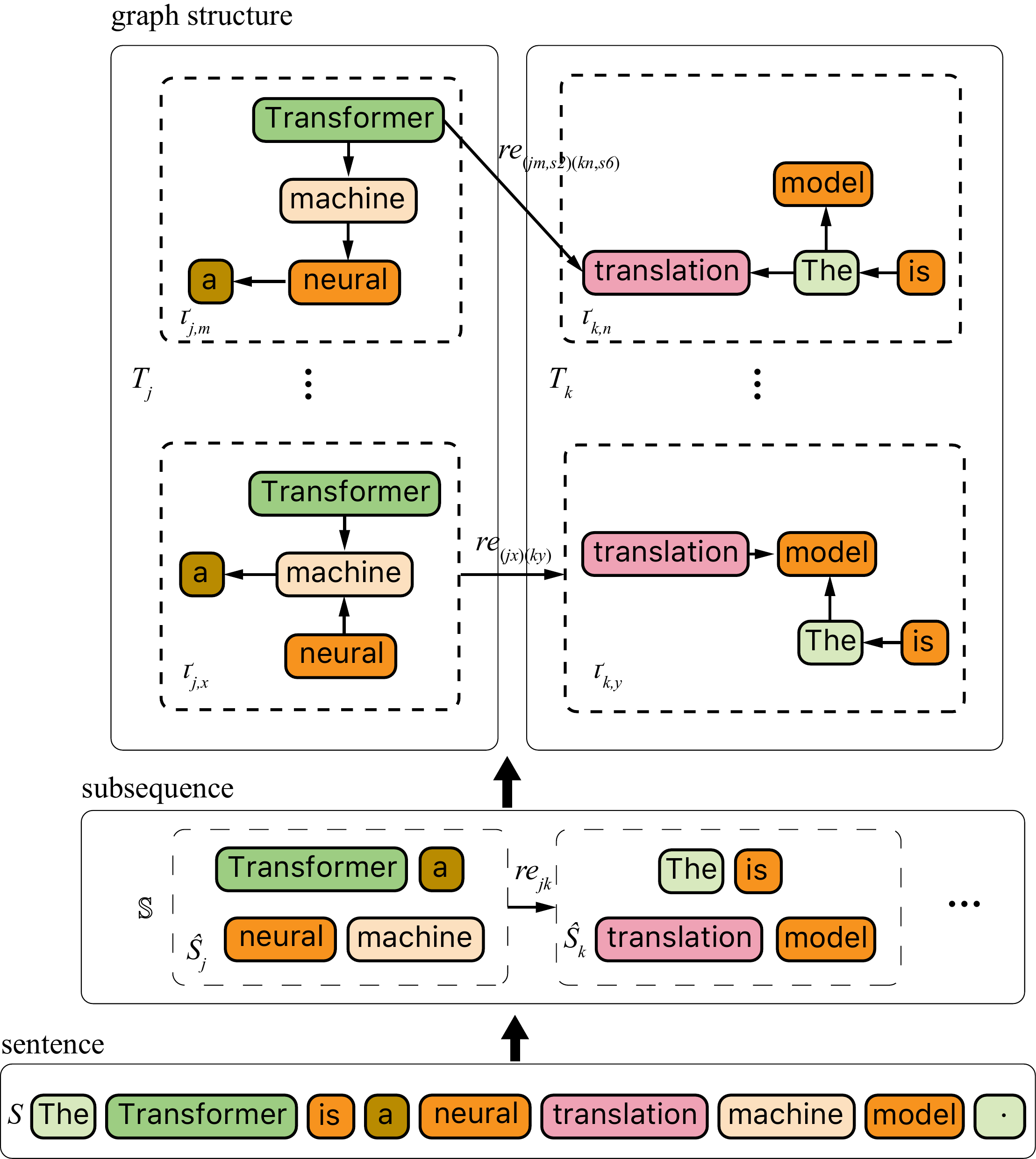}
    \caption{An example of sentence, subsequences, graph structures, and relationships. A sentence contains diverse subsequences, and encoding is a process to capture relationships between subsequences. One subsequence can be represented by different graph structures like dependency parsing tree, in which the words are vertexes and relationships between words are edges. The relationship between subsequences can be replaced by the sum of the relationships between different graph structures of respective subsequences. Furthermore, the relationship between graph structures can be replaced by the sum of the relationships between words from different graph structures.}
    \label{fig:structure}
\end{figure}

In the case of NLP, the most popular encoding objective is sentence, which consists of uncertain number of words. Researchers have reached an agreement that a sentence cannot be simply encoded or represented only by linearly combining words, even though any sentence is linearly written word by word. Actually, word relation among the same sentence must not be linear, and sentence semantics sometimes have to be derived by subsequences (i.e., multiple neighbored words) inside a sentence. All these indicate that complicated latent structures hide behind sentence encoding. To formally represent such structures, we may conveniently regard words as vertexes and the relationships among words as edges to build graphs in mathematics. Taking sentence encoding as our case study, we give a modeling explanation for the encoding and representation of the SAN-based model. Fig \ref{fig:structure} shows an example of our explanation of sentence, subsequences, graph structures, and relationships. Based on self-attention and layer-based model design, the SAN-based model can generate these various graph structures among words using previously generated structures recursively, which enables the SAN-based model to capture structural information or linguistic information (i.e., linguistic knowledge) from input sentences.

We can model the encoding into a single directed graph in which all the processed intermediate representations and the respective processing of representations are vertexes and edges respectively, and explain the encoding as a process to capture relationships between representations. Although this  modeling method is simple to understand, it is insufficient to model the complicated relationship among the original representations and deeper processing in the encoder. Intuitively, the simple directed graph can primarily models relationships between words but is incapable of modeling the relationship among phrases or clauses, which also consist of structural information. Therefore, the simple directed graph cannot describe our explanation for the ability to extract structural or linguistic information.

To describe our explanation with mathematical notation, we model the representations as a modified multigraph called \textit{Multi-order-Graph} (MoG), which primarily allows subgraphs with order specifications to model the relationship between different representations. An MoG uses vertex, edge, and subgraph to reflect word, relationship, and structure captured by the SAN-based model respectively, and the generation of MoG represents the encoding process of the SAN-based model. With MoG, we can comprehensively observe every step of the representation generation, and unify various complicated relationships inside encoders into a consistent graph. Compared with modeling the representations as a simple directed graph, MoG can not only model relationships among words but also model the relationship among phrases or clauses especially structural information, and can also model the relationship among the original representations and encoding process in the model. In addition, our MoG explanation can precisely depict model with different depths and sentences with different lengths.

Based on our explanation and proposed MoG, we analyze the current SAN encoding mechanism and propose a Graph-Transformer model which can empower the performance on sequence-to-sequence (seq2seq) tasks by further enhancing the multiple related representation encoding in terms of the proposed subgraphs from MoG. For the representation, we consider their layer-based processing procedures by distinguishing them into three parts representing subgraphs of high order, middle order, and low order in MoG. We only calculate the parts for high-order and middle-order subgraphs by self-attention to avoid generating the repeat subgraphs and focus on subgraphs of higher orders related to more complex relationships among words. We also give three methods to fuse three parts of information. We evaluate our model on IWSLT14 German-English, WMT14 English-German, WMT14 English-French, and WMT16 English-Romanian tasks, and the experimental results show that our model can improve the performance on seq2seq tasks.

The contribution of this paper is three-fold:
\begin{itemize}
    \item We propose a novel explanation for encoding of SAN-based model in which the SAN-based model captures graph structures recursively based on self-attention and layer-based structure. It can explain the ability to capture structural or linguistic information from input sentences. 
    \item We propose a modeling method in which the representations are modeled as a modified multigraph called \textit{Multi-order-Graph} (MoG) to describe our explanation. Compared with simple directed graph, MoG can model the relationship among phrases or clauses, especially structural information.
    \item We propose an enhanced Transformer called Graph-Transformer by analyzing the SAN-based model with our explanation and MoG. Our model can improve the performance on seq2seq tasks by balancing the weights of subgraphs of different orders.
\end{itemize}

\section{Related Work}
\subsection{SAN-based Model}

NMT models based on RNN\cite{kalchbrenner2013recurrent, DBLP:journals/corr/BahdanauCB14, DBLP:conf/nips/SutskeverVL14,DBLP:conf/emnlp/LuongPM15}, CNN\cite{DBLP:conf/icml/GehringAGYD17}, or SAN\cite{DBLP:conf/nips/VaswaniSPUJGKP17,shaw-etal-2018-self,DBLP:conf/nips/HeTXHQ0L18} with an encoder-decoder framework have achieved further performance improvement on several datasets. As the state-of-the-art NMT model, the Transformer is the first SAN-based model using scaled dot-product attention. Each encoder layer contains a self-attention sub-layer and a feed-forward sub-layer, while each decoder layer has one more encoder-decoder attention sub-layer following the self-attention sub-layer. The Transformer also employs a residual connection \cite{DBLP:conf/cvpr/HeZRS16} around each sub-layer followed by layer normalization \cite{ba2016layer}. Without recurrence for sequence order, the Transformer uses position encoding \cite{DBLP:conf/icml/GehringAGYD17} to mark the position. To ensure that the predictions cannot use unknown outputs, the self-attention sub-layer prevents positions from attending to subsequent positions with a mask.

Several variants have been proposed to improve the performance of the original Transformer. Shaw et al.\cite{shaw-etal-2018-self} proposed relative position representations in the self-attention mechanism to replace the absolute position encoding and enhance the ability to capture local information of the input sentence. He et al.\cite{DBLP:conf/nips/HeTXHQ0L18} shared the parameters of each layer between the encoder and decoder to coordinate the learning between encoder and decoder. Yang et al.\cite{DBLP:conf/emnlp/YangTWMCZ18} proposed a model which enhances the ability to capture useful local context by casting localness modeling as a learnable Gaussian bias and improves the performance on two translation tasks. Wang et al.\cite{DBLP:conf/acl/WangLXZLWC19} worked on the proper use of layer normalization and a novel way of passing the combination of previous layers to the next, and train a 30-layer encoder that outperforms Transformer-Big on some tasks. Zhang et al.\cite{DBLP:conf/naacl/ZhangLFZ19} presented an implicit syntax encoding method for NMT by syntax-aware word representations.  You et al.\cite{DBLP:conf/acl/YouSI20} improved memory efficiency and decoding speed without significantly lowering BLEU with hard-coded attention. Raganato et al.\cite{DBLP:conf/emnlp/RaganatoST20} replaced all but one attention head of each encoder layer with simple fixed non-learnable attentive patterns and increased BLEU scores in low-resource scenarios. Dehghani et al.\cite{DBLP:conf/iclr/DehghaniGVUK19} proposed Universal Transformers, which uses a dynamic per-position halting mechanism to choose the required number of refinement steps for each symbol and improves accuracy on several tasks. Kitaev et al.\cite{DBLP:conf/iclr/KitaevKL20} replaced dot-product attention with locality-sensitive hashing attention and added reversible layers in the proposed Reformer, which is much more memory-efficient and faster on long sequences. Gu et al.\cite{DBLP:conf/iclr/Gu0XLS18} avoided autoregressive decoding, predicted outputs in parallel and achieved a near-state-of-the-art performance on some NMT tasks.

SAN-based model is also used for other NLP tasks. Li et al.\cite{li-etal-2020-flat} proposed Flat-Lattice Transformer for Chinese NER and outperformed other lexicon-based models in performance and efficiency. Kitaev and Klein\cite{kitaev-klein-2018-constituency} replaced an RNN encoder with a self-attentive architecture to improve the performance of the constituency parser. Koncel-Kedziorski et al.\citet{koncel-kedziorski-etal-2019-text} proposed a Graph-based model for text generation to extend the successful Transformer for text encoding to graph-structured inputs and incorporate global structural information. BERT, proposed by Devlin et al.\cite{DBLP:conf/naacl/DevlinCLT19}, is a widely-used pre-trained language model based on the SAN-based model, and improves the performance on various NLP tasks, especially natural language understanding tasks\cite{wang2019glue}. 
BERT allows researchers to use representation from existing language models and simply fine-tune all pre-trained parameters to train the model on the downstream tasks, which makes well-designed pre-trained language models popular in NLP tasks. Dai et al.\cite{dai-etal-2019-transformer} enabled the SAN-based language model to learn dependency beyond fixed length without disrupting temporal coherence. RoBERTa proposed by Liu et al.\cite{DBLP:journals/corr/abs-1907-11692} exceeds the performance of BERT by some modifications such as training the model longer with bigger batches over more data. Lan et al.\cite{DBLP:conf/iclr/LanCGGSS20} propose ALBERT to achieve better performance than BERT-large using fewer parameters by two parameter reduction techniques. The SAN-based language model also improves the performance on NMT tasks.  Zhu et al.\cite{Zhu2020Incorporating} used BERT as another embedding to extract representations for input sentences and fuse the representation with the  encoder and decoder through attention mechanisms. Yang et al.\cite{Yang_Wang_Zhou_Zhao_Zhang_Yu_Li_2020} proposed the concerted training approach to make the most use of BERT in NMT. Xu et al.\cite{DBLP:conf/emnlp/XuDM21} proposed BIBERT for English-German NMT and dual-directional translation models, which leverages the inherent bilingual nature of BIBERT with mixed domain training and fine-tuning. BART\cite{DBLP:conf/acl/LewisLGGMLSZ20} is a denoising autoencoder for pre-training seq2seq models which achieves new state-of-the-art results and makes a 1.1 BLEU score  increase on the NMT task. Guo et al.\cite{DBLP:conf/nips/GuoZXWCC20} proposed a flexible and efficient model, which is able to jointly leverage the information contained in the source-side and target-side BERT models and outperforms autoregressive baselines.

\subsection{Structural and Linguistic Information Learning in SAN-based Model}
\label{section:structure}
Existing works on RNN-based models show that RNN-based models can learn syntactic information and structure from data. Following these works, researchers found that SAN-based models can also extract structure information and linguistics knowledge. Shi et al.\cite{DBLP:conf/emnlp/ShiPK16} proposed two methods to find that different syntactic information tends to be stored at different layers in the NMT models. Blevins et al.\cite{DBLP:conf/acl/BlevinsLZ18} found a correspondence between network depth and syntactic depth, suggesting that a soft syntactic hierarchy emerges. Belinkov et al.\cite{DBLP:conf/acl/BelinkovDDSG17} evaluated the NMT model on two tasks and showed that different representations are captured in different layers of the model, and the target language impacts the kind of information from the model. Niven and Kao.\cite{DBLP:conf/acl/NivenK19} evaluated BERT on the argument reasoning comprehension task and claimed that BERT has learned nothing about argument comprehension while BERT is indeed  a powerful learner. Jawahar et al.\cite{DBLP:conf/acl/JawaharSS19} showed that layers at the top, middle and bottom could capture semantic, syntactic, and surface features from sentences. It means that BERT can learn some linguistics information from data. Kovaleva et al.\cite{DBLP:conf/emnlp/KovalevaRRR19} proposed a methodology and carried out a qualitative and quantitative analysis of the information encoded by the individual heads of BERT and showed that manually disabling attention in certain heads improves  the performance of BERT. Marecek and Rosa\cite{DBLP:conf/emnlp/MarecekR18} analyzed the encoder in English-to-German NMT and proposed  algorithms for constructing syntactic trees. Goenen et al.\cite{DBLP:conf/nips/ReifYWVCPK19} showed the evidence of syntactic representation in attention matrices. Tran et al.\cite{DBLP:conf/emnlp/TranBM18} compared LSTM with Transformer in the ability to capture the underlying hierarchical structure of sequential data and showed that LSTMs slightly but consistently outperform the Transformer. They also showed that LSTMs generalize better than the Transformer to longer sequences in a logical inference task. Miaschie et al.\cite{DBLP:conf/coling/MiaschiBDV20} found that BERT can encode a wide range of linguistic characteristics. Vig and Belinkov\cite{DBLP:conf/blackboxnlp/VigB19} showed that self-attention in language model captured  different relationships  at different layer depths. Hahn\cite{DBLP:journals/tacl/Hahn20} showed that self-attention could not model periodic finite-state languages or hierarchical structures  unless the number of layers or heads increases with input length. 

The SAN-based model indeed extracts the structural or linguistic information, especially syntactic features and semantic features, from input sentences. However, the works above found some characteristics of SAN-based models: 
\begin{itemize}
    \item information captured by different layers in the SAN-based models are different, and deeper layers often capture more complex features, such as syntactic features in the middle and semantic features at the top;
    \item the model depth can influence the performance that adding layers to the model may improve the performance while too many layers may hurt the performance, and different tasks require different model depths;
    \item to get the best performance, sentences with different lengths require model depths.
\end{itemize}
We wonder how the SAN-based model extracts the structural or linguistic information from input sentences and the relationship between performance and the model architecture. Following existing works and their results, we give an explanation for the characteristics above in Section \ref{section:relations}.

\section{Proposed Explanation and Modeling Method}

In some NLP tasks, such as dependency parsing, we can build a tree by connecting generated sub-tree. 
Inspired by tree-building algorithms in dependency parsing tasks and related works in Section \ref{section:structure}, we explain the SAN-based model encoding as a capturing process of the information of words and relationships among them, which can be represented by a graph structure of subsequence. It means that the SAN-based model encoding can be explained as an iterative process to build new edges to connect previously generated graph structures and get new graph structures instead of building every new structure from the roots. We describe our explanation in detail in Section \ref{section:relations}. Following our explanation, we introduce our Multi-order-Graph (MoG) in Section \ref{subsection:multigraph}.

\subsection{Explanation for Encoding of SAN-based Model}
\label{section:relations}
We argue that a sentence is a set of words and relationships between every two subsequences, which can be phrases, subordinate clauses, and compound words. Information extracted from the sentence is not only word information but also relationship information. Relationship information extracted from the sentence can be replaced by relationships between graph structures of different subsequences.

Let us start with the definition of subsequences, graph structures, and relationships among graph structures. We define $\hat{S}_j$ is the $j$-th subsequence of sentence $S=(s_1,...,s_n)$ where $1 \leq j \leq 2^n-1$.  $\hat{S}_j$ is a word sequence like $S$ and has $2^{len(\hat{S}_j)}-1$ subsequences where $len(\hat{S}_j)$ is the length of $\hat{S}_j$. Note that all subsequences of $\hat{S}_j$ are also subsequences of $S$. We define $re_{jk}$ as the relationship between $\hat{S}_j$ and $\hat{S}_k$.

Several possible graph structures like the dependency parsing tree can be used to represent one subsequence. We use $\tau_{j,x}$ for the $x$-th graph structure of $\hat{S}_j$. We define $re_{(j,x)(k,y)}$ as the relationship between $\tau_{j,x}$ and $\tau_{k,y}$.

Given graph structures $\tau_{j,x}$ and $\tau_{k,y}$, we define $re_{(j,x,s_a)(k,y,s_b)}$ for the relationship between words $s_a$ and $s_b$ where $s_a$ and $s_b$ belong to ${\tau_{j,x}}$ and ${\tau}_{k,y}$ respectively, and  we call $re_{(j,x,s_a)(k,y,s_b)}$ \textbf{dependency relationship} between $s_a$ and $s_b$.  It is easy to know that relationships between the same two words may change in different sentences. More generally, dependency relationships between words can change when words belong to different structure pairs, and we can get  
\begin{equation}
    re_{(j,x,s_a)(k,y,s_b)} \neq re_{(l,z,s_a)(m,w,s_b)},
    \label{p2}
\end{equation}
where $<j,k,x,y> \neq <l,m,z,w>$.

Inspired by tree-building algorithms in dependency parsing tasks, we explain that the graph structure uses words and dependency relationships between words as vertexes and edges, and is built recursively in which graph structures having only one word are generated first and then other graph structures are generated by building dependency relationships between two words from two generated graph structures. It means that graph structures cannot be generated in random order, and new graph structures with more words are built using graph structures with fewer words. Note that one graph structure will not be removed if it is used to generate other graph structures. 

Given $\tau_{j,x}$ built by using $re_{(k,y,s_a)(l,z,s_b)}$, because building $re_{(k,y,s_a)(l,z,s_b)}$ generates $\tau_{j,x}$ and is determined by $s_a$, $s_b$, $\tau_{k,y}$ and $\tau_{l,z}$, one dependency relationship $re_{(k,y,s_a)(l,z,s_b)}$ can be regarded as a part of relationship $re_{(k,y)(l,z)}$ between $\tau_{k,y}$ and $\tau_{l,z}$ or one relationship among all words from $\tau_{j,x}$.

We explain the encoding of $S$ as a process to capture all $re_{jk}$ rather than only word-wise relationships in which the model calculates dependency relationships to replace $re_{jk}$ approximately instead of calculating $re_{jk}$ directly. Given $re_{(j,x)(k,y)}$, we can replace the relationship as the sum of dependency relationships between words in $\tau_{j,x}$ and words in $\tau_{k,y}$ as 
\begin{equation}
    re_{(j,x)(k,y)}\approx \sum_a \sum_b re_{(j,x,s_a)(k,y,s_b)}.
    \label{p3}
\end{equation}
Furthermore, the relationship $re_{jk}$ can also be approximately replaced with the sum of relationships between structures of subsequences such as
\begin{equation}
    re_{jk} \approx \sum_x \sum_y re_{(j,x)(k,y)}
    \label{p1}
\end{equation}
Therefore, we can use the sum of dependency relationships as 
\begin{equation}
    re_{jk} \approx \sum_x\sum_y(\sum_a\sum_b re_{(j,x,s_a)(k,y,s_b)}) 
\end{equation}
to replace the relationships $re_{jk}$ between $\hat{S}_j$ and $\hat{S}_k$.

Based on the discussion about graph structures and dependency relationships above, we explain that the SAN-based model builds graph structures and captures dependency relationships recursively which enables the SAN-based model to capture structural information. In the SAN-based model, a layer uses representations from the previous layers, i.e., information of the captured graph structures, to calculate dependency relationships among words, represented by values in self-attention matrices, and generate new representations, a set of new graph structures. Self-attention allows the word to access other words and the model to connect every two words ignoring their distances and orders, which makes the SAN-based model build dependency relationships between every two words and generate various graph structures. Besides, the graph structures built by the SAN-based model having the same topology as linguistic structures such as the dependency parsing tree and semantic role labeling tree can be regarded as linguistic information from the input sentence.

In the SAN-based model, the deeper layers capture more complex graph structures than the lower layer. The maximum number of words in a structure generated by SAN layers grows exponentially with the depth increasing. As the SAN-based model layers can only use representation from the previous layers, if the $i$-th layer builds graph structures with no more than $m$ words, the $(i+1)$-th layer can only build graph structures with no more than $2m$ words. Thus, the first layer captures structures containing two words, and the $i$-th layer captures structures with no more than $2^i$ words. It answers why different layers in the SAN-based model capture different structural or linguistic information.

Therefore, given a model with $m$ layers and a sentence with length $l$, the model can capture all structures in the first $\mathrm{log}_2({l})$ layers ($\mathrm{log}_2(l) \le m$), and the model cannot capture structures among entire sentences if the sentence length exceeds $2^m$. For the sentence with $l$ words, the model needs $\mathrm{log}_2({l})$ layers to generate graph structures among all words. Thus, sentences with different lengths require different depths of models for the best performance. For the long sentence, the model performance is limited because the model cannot capture graph structures from enough words.

Because all graph structures are built by connecting captured structures, the model may build redundant structures such as connecting identical structures. Residual connections in the SAN-based model incorporate representations from the previous layers into representations generated by the current layer, which allows the following layers to build redundant structures. If with too many layers, the model can generate many redundant structures and hurt the performance.

Though the SAN-based model can capture linguistic information from the sentence, the key to the SAN-based model is that the self-attention mechanism and layer architecture allow the model to build various graph structures among words iteratively.
The graph structures represent different kinds of information such as phrases, clauses, and even relationships, e.g., the word You is often followed by the word were in the simple past tense. Therefore, we should not only focus on known linguistic information captured by the SAN-based model but also pay more attention to unknown linguistic information.

Our explanation can also be extended to other models.
For example,  the encoding in the RNN-based model can be regarded as a process to generate various graph structures. Unlike the SAN-based model in which the current layer can only use graph structures generated in the previous layers, the layer in the RNN-based model can use graph structures generated in the previous time steps to build new graph structures.

\subsection{Definition of Multi-order-Graph}
\label{subsection:multigraph}

\begin{figure}[t]
\centering
\subfigure[Different edges between \textit{machine} and \textit{translation} when words belong to different subgraphs.]{
\begin{minipage}[t]{1\linewidth}
\centering
\includegraphics[scale=0.125]{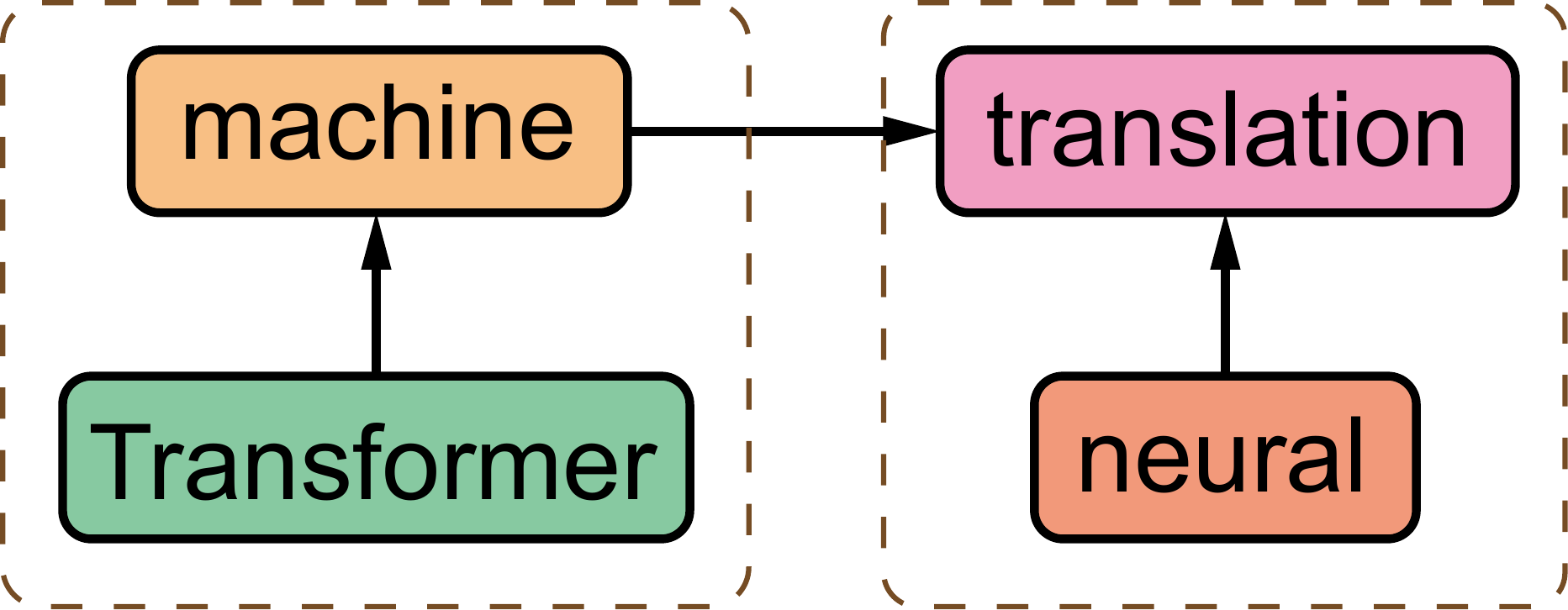}
\includegraphics[scale=0.125]{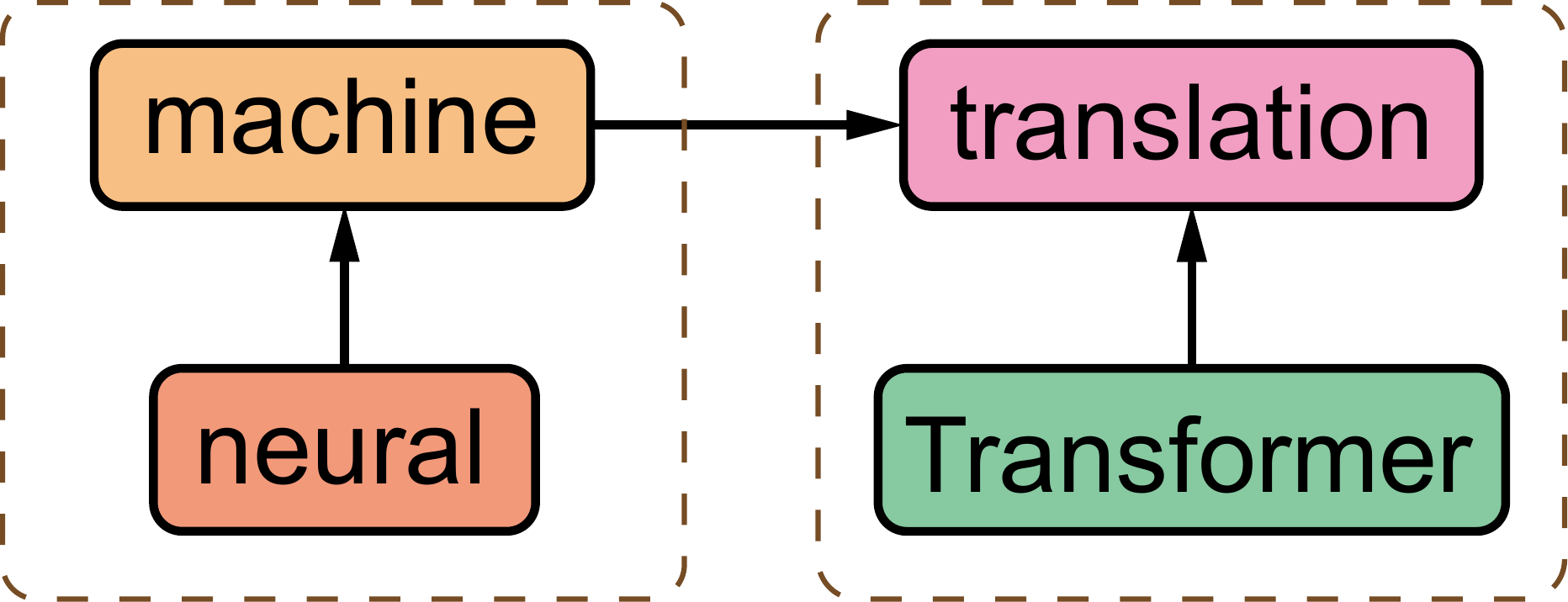}
\label{figure:dependency1}
\end{minipage}
}

\subfigure[Different edges between \textit{machine} and \textit{translation} with different directions]{
\begin{minipage}[t]{1\linewidth}
\centering
\includegraphics[scale=0.125]{figure/dependency1.pdf}
\includegraphics[scale=0.125]{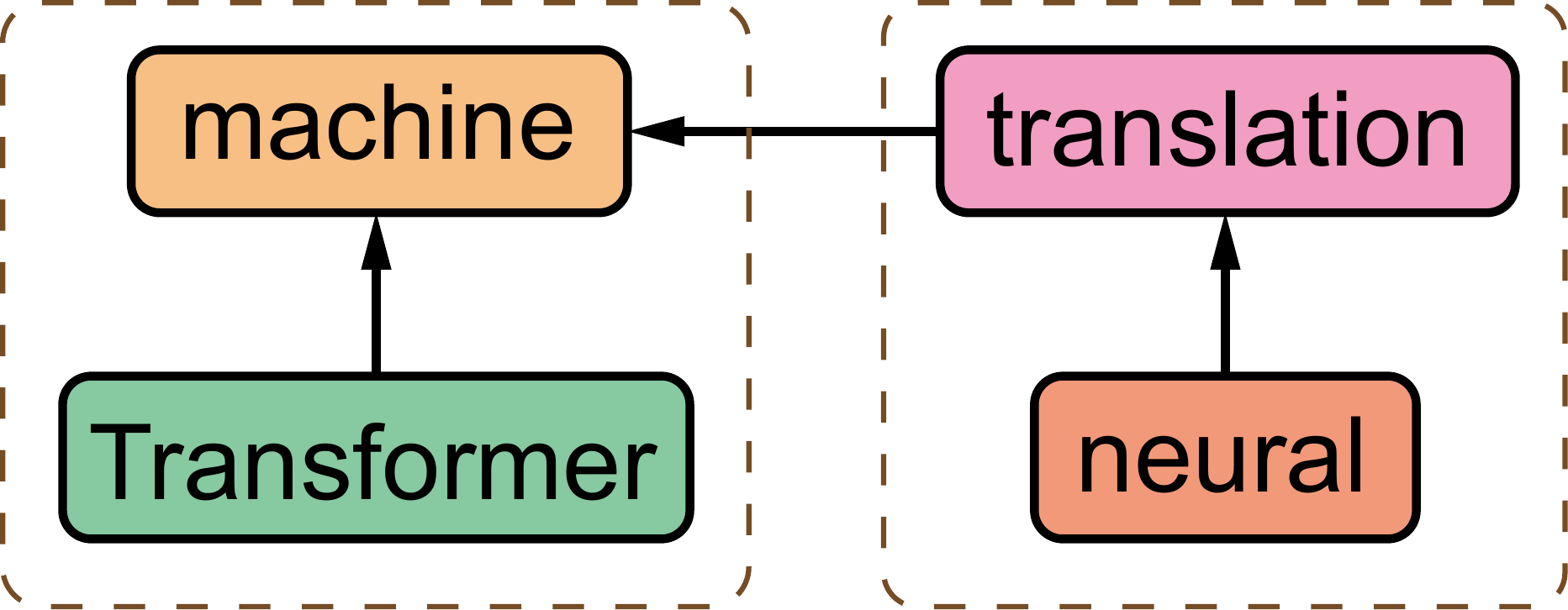}
\label{figure:dependency5}
\end{minipage}
}
\\
\subfigure[Different edges between \textit{machine} and \textit{translation} when \textit{machine} belongs to different subgraphs.]{
\begin{minipage}[t]{1\linewidth}
\centering
\includegraphics[scale=0.125]{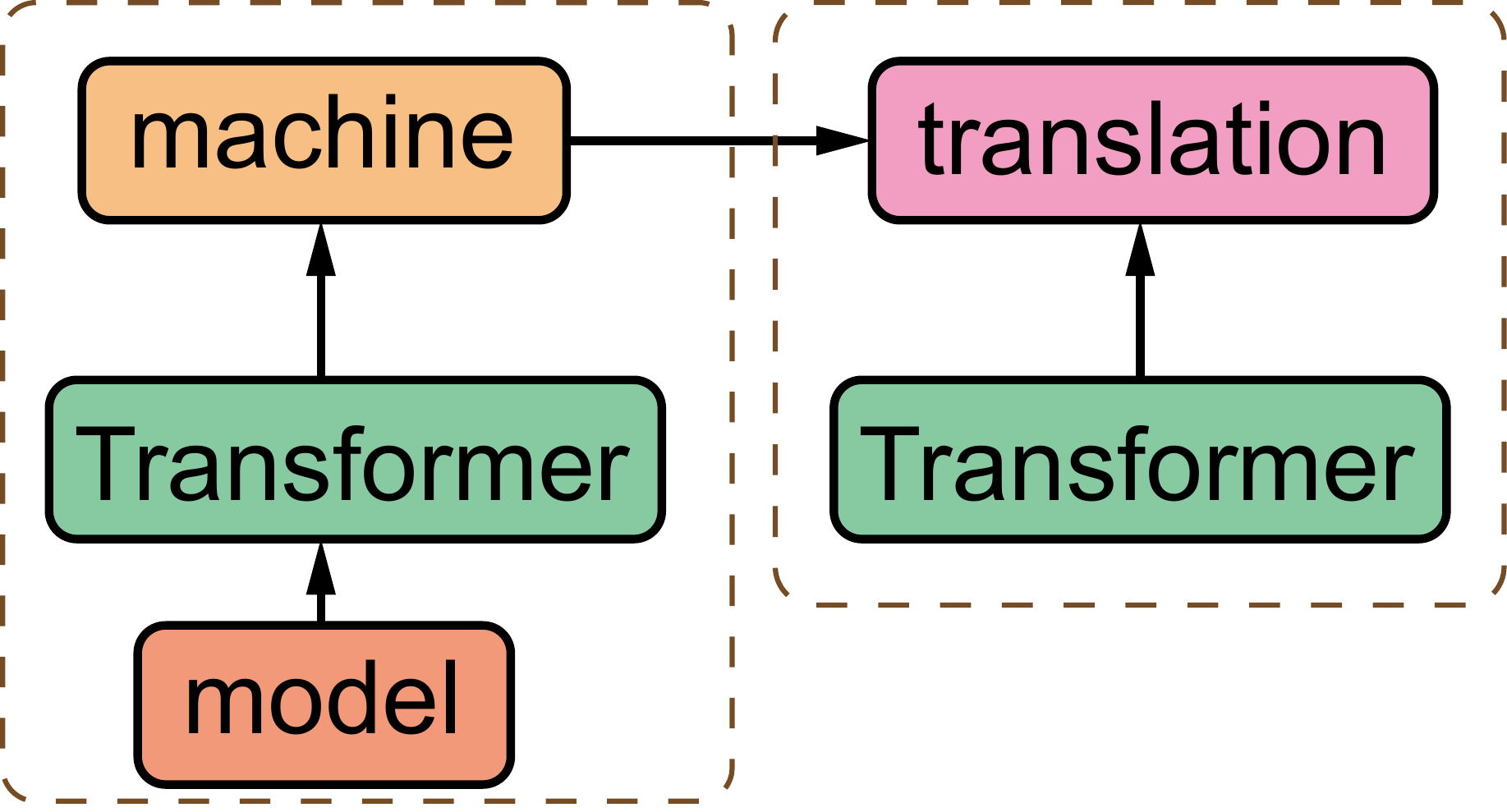}
\includegraphics[scale=0.125]{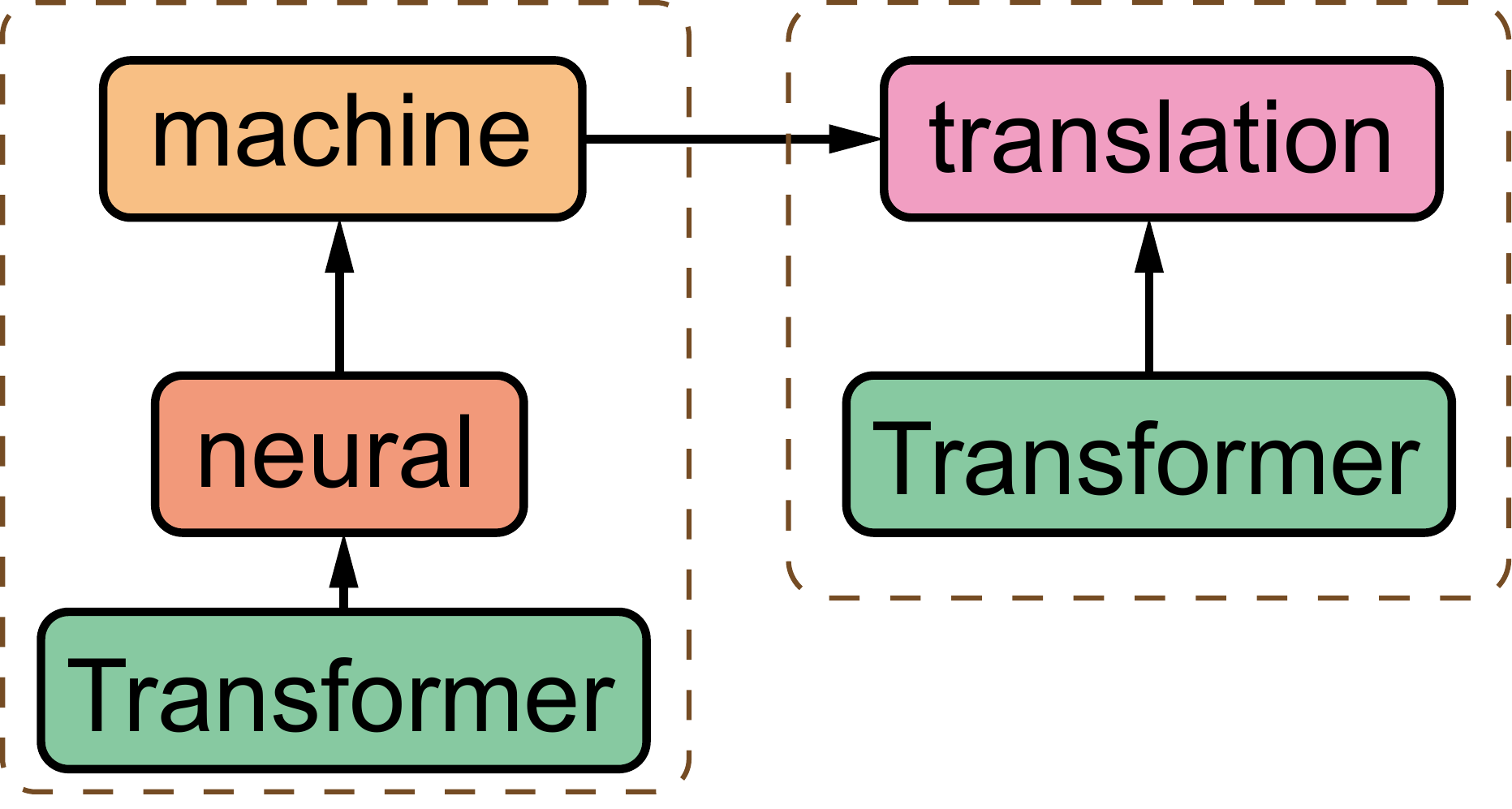}
\label{figure:dependency6}
\end{minipage}
}

\caption{Given sentence \textit{The Transformer is a neural machine translation model.}, these figures show relationships between word \textit{machine} and \textit{translation}. Two subgraphs  in each group have the same topology.}
\label{figure:dependency}
\end{figure}

A simple directed graph that uses vertexes for words and edges for relationships between words can reflect the representation of a sentence. However, it is difficult to represent graph structures because there are multiple edges between the same two vertexes for various structures, while a simple directed graph has only one edge between two vertexes. Furthermore, a simple directed graph is also tough to record the generation of graph structures.

In graph theory, a directed multigraph (or pseudograph) is a graph in which more than one directed edge connects two vertexes. In this section, we introduce a multigraph called Multi-order-Graph (MoG) for representation of the input, which defines edges to reflect the relationship between representations more comprehensively.

Intuitively, an MoG is an extended multigraph in which an edge connects not only vertexes but also subgraphs containing these vertexes. Formally, we define MoG as a tuple $G=(V^G, E^G)$, where $V^G=\{v^G_1,...,v^G_n\}$ is a finite set of vertexes and $E^G=\{e^G_1,...,e^G_m\}$ is the finite set of edges. For edge $e^G_k$, the source and target vertexes are $SN(e^G_k)$ and $TN(e^G_k)$ where $SN(\cdot)$ and $TN(\cdot)$ are functions to map edges to source and target vertexes respectively.  In MoG, we use vertex, edge, and subgraph for word, dependency relationship, and graph structure of sentence respectively, and the subgraph and edge are the base and the distinct parts of MoG.

A subgraph of $G$ is a graph with no less than a word defined as $sub^G_j=(V^G_j,E^G_j)$, in which $V^G_j \subseteq V^G$, $E^G_j \subseteq E^G$. The order of $sub^G_j$ is the number of vertexes in $sub^G_j$ and equal to $|V^G_j|$, which can be defined as $order_s(sub^G_j)$. The simplest subgraph has one vertex and no edge, and its order is 1. We define $Sub^G=\{sub^G_1,...,sub^G_p\}$ as the set of all subgraphs of $G$. In MoG, a subgraph is related to a graph structure of sentence and has the same topology as the graph structure. Fig. \ref{figure:dependency} shows some subgraphs of the sentence \textit{The Transformer is a neural machine translation model}.

\begin{figure*}[!htbp]
\centering
\subfigure[No loop or overlap.]{
\begin{minipage}[t]{0.22\linewidth}
\centering
\includegraphics[scale=0.25]{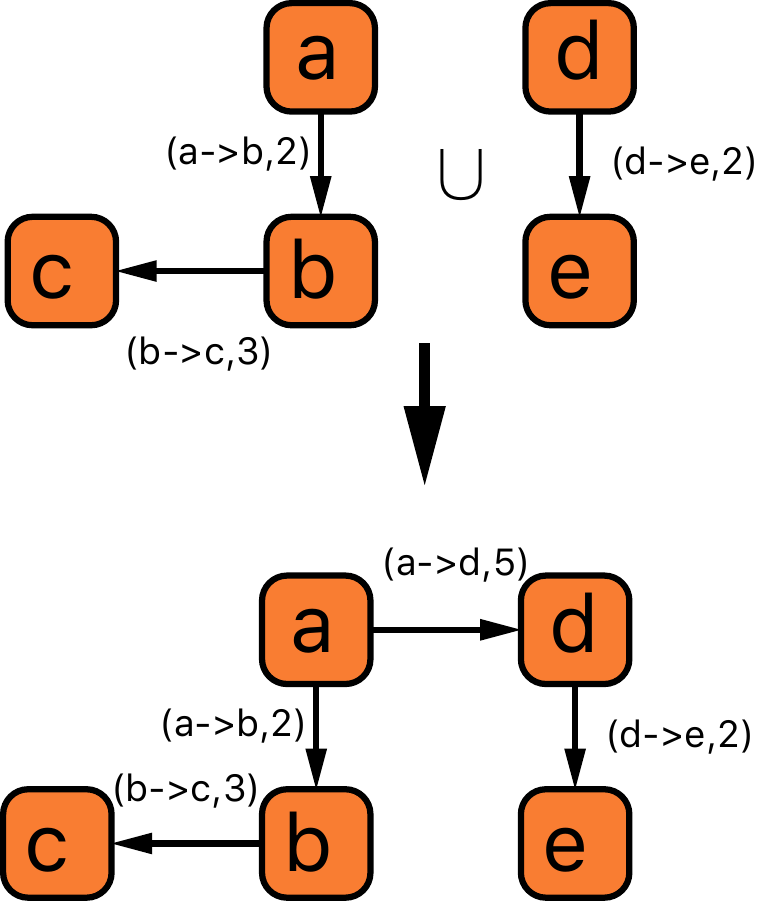}
\label{figure:acyc}
\end{minipage}
}
\subfigure[Loop.]{
\begin{minipage}[t]{0.22\linewidth}
\centering
\includegraphics[scale=0.25]{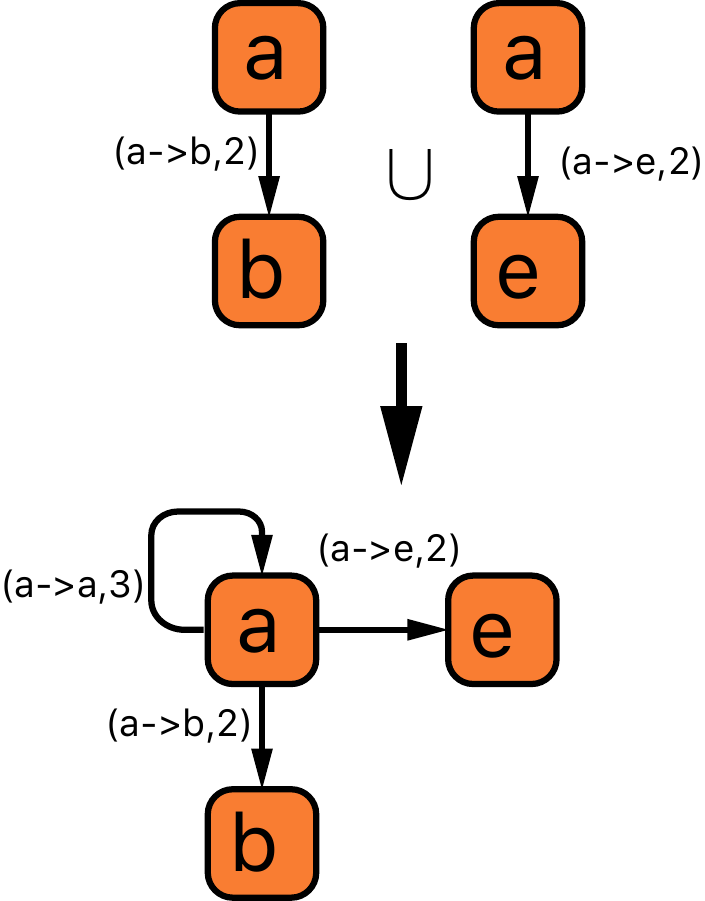}
\end{minipage}
}
\subfigure[Overlap of vertexes.]{
\begin{minipage}[t]{0.22\linewidth}
\centering
\includegraphics[scale=0.25]{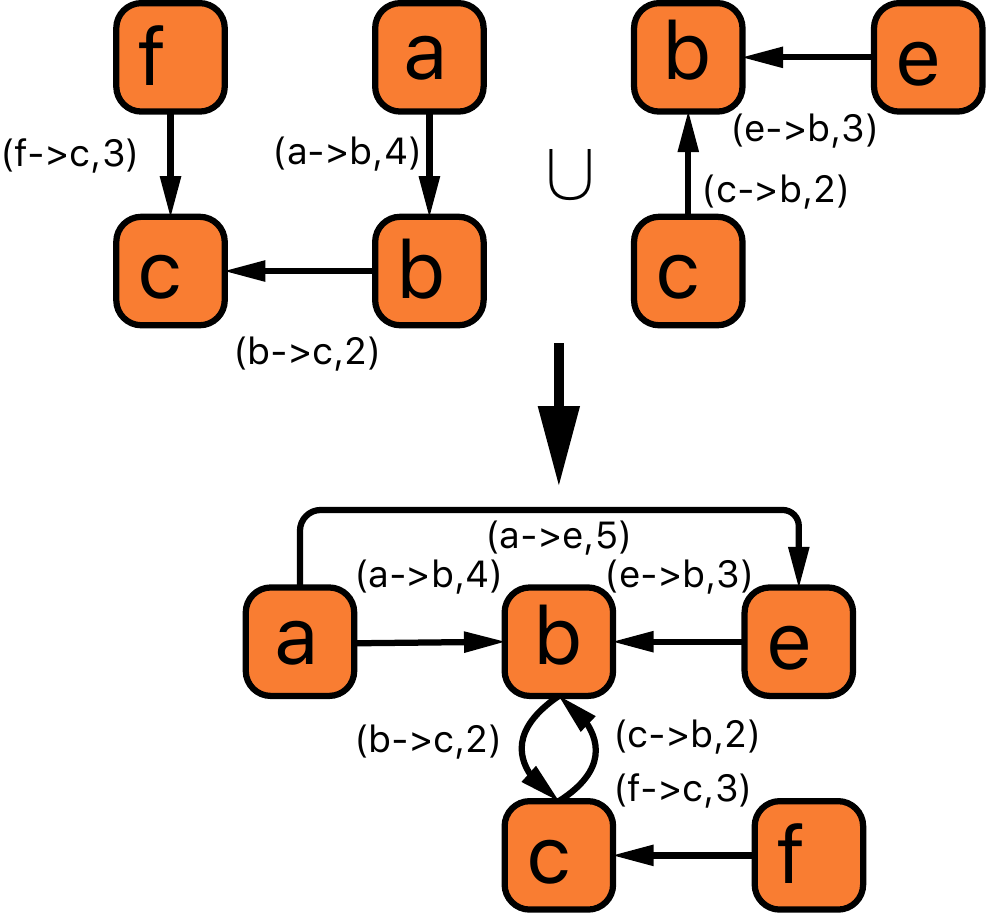}
\end{minipage}
}
\subfigure[Overlap of vertexes and edges.]{
\begin{minipage}[t]{0.22\linewidth}
\centering
\includegraphics[scale=0.25]{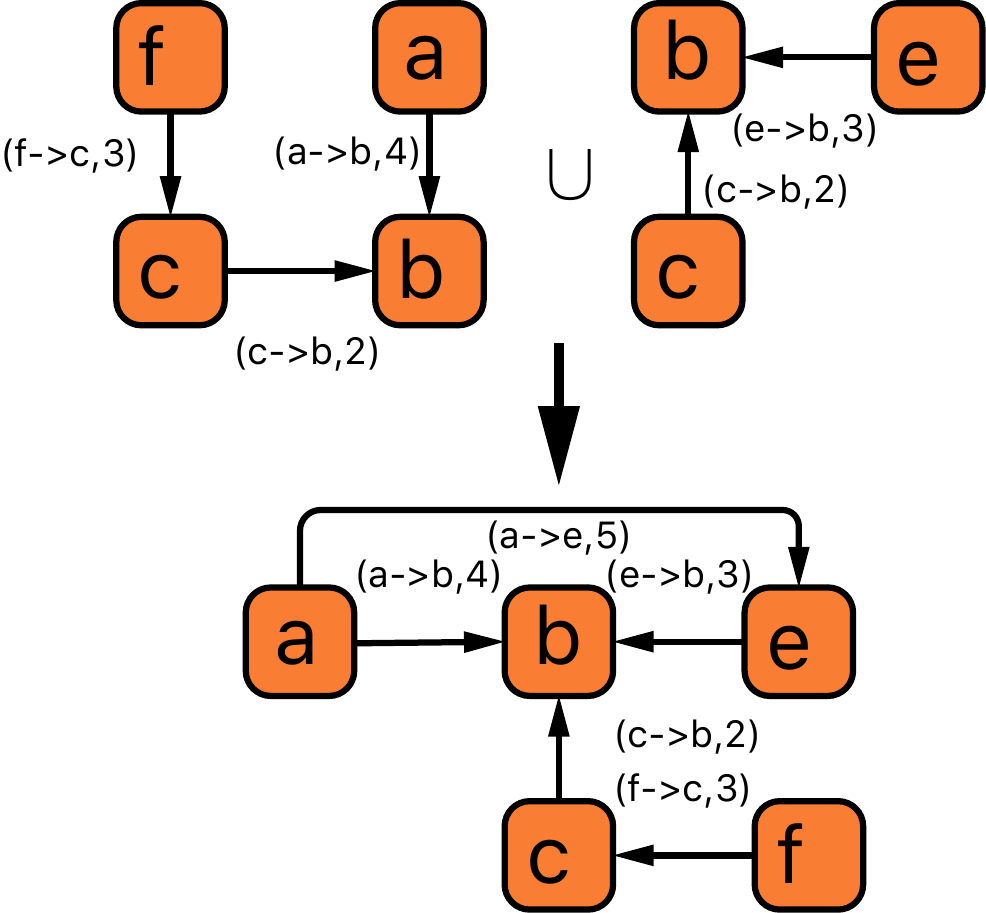}
\end{minipage}
}
\caption{Generation of different kinds of subgraph.}
\label{figure:generation}
\end{figure*}

\begin{figure}[htb]
    \centering
    \includegraphics[scale=0.27]{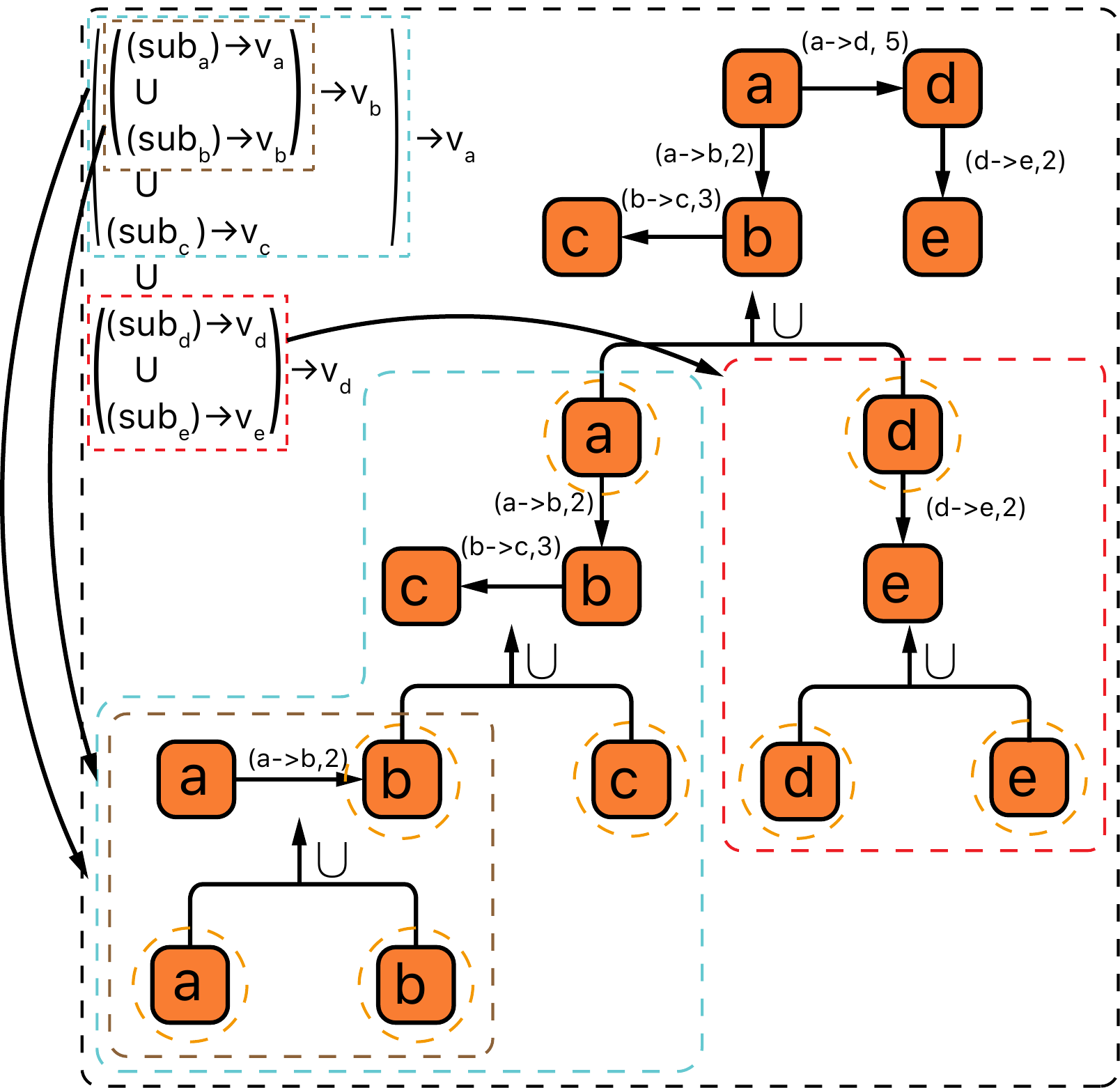}
    \caption{An example of subgraph generation tree in Eq (\ref{eq:binary}). }
    \label{fig:2tree}
\end{figure}
\begin{figure}[htb]
\flushleft
\subfigure[Edge $(a \rightarrow d,5)$.]{
\begin{minipage}[t]{0.5\linewidth}
\centering
\includegraphics[scale=0.25]{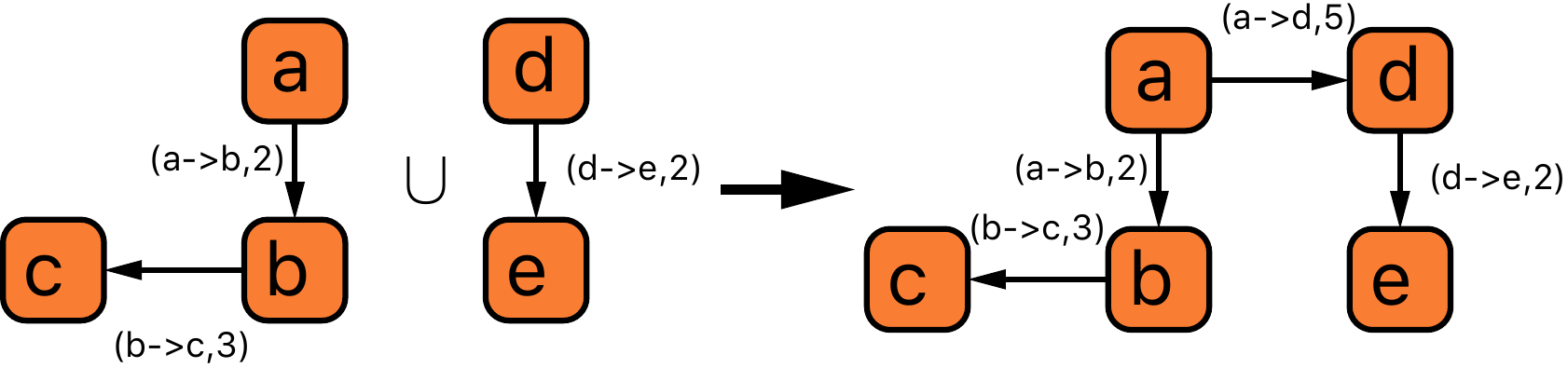}
\label{figure:subgraph1}
\end{minipage}
}
\subfigure[Edge $(a \rightarrow b,5)$.]{
\begin{minipage}[t]{0.5\linewidth}
\centering
\includegraphics[scale=0.25]{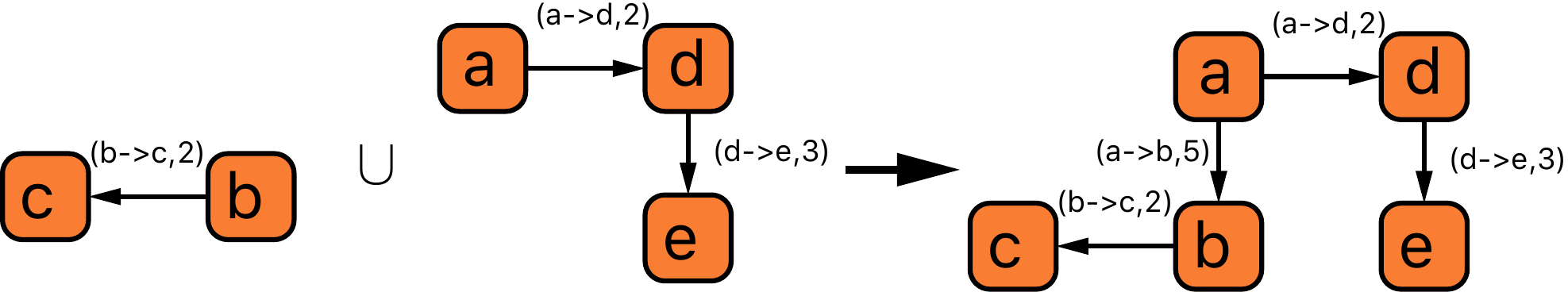}
\label{figure:subgraph2}
\end{minipage}
}
\caption{Generation of two different subgraphs with the same topologies.}
\label{figure:subgraph}
\end{figure}

An edge in MoG represents one dependency relationship. According to Eq (\ref{p2}), different edges will be built between two same vertexes when vertexes belong to different subgraph pairs. It indicates that one edge $e^G_j$ connected $v^G_k \in V^G_m$ and $v^G_l \in V^G_n$ is determined by four variables:
\begin{itemize}
\item source vertex $v_k^G$ of edge $e^G_j$,
\item target vertex $v^G_l$ of edge $e^G_j$,
\item source subgraph $sub^G_m$ in which $v_k^G \in V^G_m $,
\item target subgraph $sub_n^G$ in which $v_l^G \in V^G_n$.
\end{itemize}
The edge $e^G_j$ can also connect $sub^G_m$ and $sub^G_n$ because building a new edge will generate a new subgraph, which reflects the generation of graph structures.

Building $e^G_j$ can generate a new subgraph which is a combination of $e^G_j$, $sub^G_m$, and $sub^G_n$, and it reflects the generation of graph structures. We call the new subgraph \textbf{related subgraph} of $e^G_j$ and use $sub^G_{R(j)}$ to represent it, where $R(j)$ is a function to get the identifier of the related subgraph of $e^G_j$. Setting $k=R(j)$, we call $e^G_j$ the \textbf{related edge} of $sub^G_k$, and $R^{-1}(k)$ is the inverse function of $R(j)$ to get the identifier of the related edge of $sub^G_k$. To reflect the importance of $e^G_j$ and the complexity of $sub^{G}_{R(j)}$, we define the order of $e^G_j$, denoted by $order_e(e^G_j)$ and equal to  $order_s(sub^{G}_{R(j)})$. We can use a 6-tuple 
\begin{equation}
e^G_j=(v^G_k, v^G_l, sub^G_m, sub^G_n, sub^G_{R(j)}, order_e(e^G_j))
\label{eq:edge}
\end{equation}
to present edge $e^G_j$, where $v^G_k \in V^G_m$ and $v^G_l \in V^G_n$. If we only focus on the source and target vertices, we use $(v_k^G \rightarrow v^G_l, order_e(e^G_j))$ for $e^G_j$. Note that subgraphs with one vertex and no edge have no related edge. When presented as the 6-tuple, the one vertex is for the first variable and NULL for other variables, which is revealed to the input embedding in the SAN-based model. Given the edge $e^G_j=(v^G_k, v^G_l, sub^G_m, sub^G_n, sub^G_{R(j)}, order_e(e^G_j))$, where $sub^G_m$ and $sub^G_n$ represent graph structures $\tau_{px}$ and $\tau_{qy}$of subsequences $\hat{S}_p$ and $\hat{S}_q$, respectively, $e^G_j$ represents the dependency relationship $re_{(p,x,s_k)(q,y,s_l)}$.

The building of MoG reflects the generation of graph structure and is based on building edges and subgraphs. Fig. \ref{figure:generation} shows the generation of four kinds of subgraphs.  To understand the generation of subgraphs clearly, we only focus on subgraphs without loop and overlap, which is the most simple kind of subgraph.  Generation of MoG is a recursive process in which subgraphs having one vertex and no edge are the start point of generation and other subgraphs are built using new edges to connect vertexes of generated subgraphs. It means that subgraphs and their related edges cannot be built in random order. Note that one subgraph will not be removed if it is used to build new subgraphs.

To express subgraph generation clearly, we define formula
\begin{equation}
sub_k = (sub_i)\rightarrow v_m \cup (sub_j)\rightarrow v_n,
\label{eq:express}
\end{equation}
as the operation which builds a new edge $(v_m,v_n,sub_i,sub_j,sub_k,|V_i|+|V_j|)$ and a new subgraph $sub_k$, where $|V_i|$ and $|V_j|$ are orders of $sub_i$ and $sub_j$,  $v_m \in sub_i$ and $v_n \in sub_j$ are the source vertex and the target vertex of the new edge, and $sub_k$ is generated by connecting $sub_i$ and $sub_j$.  For example, the generation of subgraphs in Fig. \ref{figure:acyc} can be expressed as
\begin{equation}
\begin{aligned}
&(((sub_a)\rightarrow v_a \cup (sub_b)\rightarrow v_b)\rightarrow v_b \cup (sub_c)\rightarrow v_c)\rightarrow v_a\\
&\cup ((sub_d)\rightarrow v_d \cup (sub_e)\rightarrow v_e)\rightarrow v_d,
\end{aligned}
\label{eq:binary}
\end{equation}

\noindent where $sub_a$, $sub_b$, $sub_c$, $sub_d$ and $sub_e$ are subgraphs with only one vertex. Note that commutative, distributive, and associative properties do not apply in this formula.

The generation of the subgraph can be expressed as binary tree like Fig. \ref{fig:2tree} which we call it \textbf{generation tree}. The generation tree can record the process of building subgraphs and can be used to distinguish subgraphs with the same topology structure. If the generation trees of subgraphs are different, these subgraphs should be considered different even with the same topologies, such as subgraphs in Fig. \ref{figure:subgraph1} and Fig. \ref{figure:subgraph2}.

\begin{algorithm}[htb]
  \caption{$i$-th Generation Step of MoG}
  \label{alg:r1}
  \begin{algorithmic}[1]
  \Require
    \Statex $Sub^G_i=\{sub^G_1,...,sub^G_p\}$
    \Statex $G_i=\{V^G, E^{G_i}\}$
  \Ensure Output $Sub^G_{i+1}$ and $G_{i+1}$
  \State $Sub_{n},E_{n}\leftarrow \emptyset,\emptyset$
  \For{$j$ \textbf{from} $1$ \textbf{to} $p$}
    \State \textbf{if} {$order_s(sub^G_j) = |V^G|$} \textbf{then} continue \textbf{end if}
    \For{$l$ \textbf{from} $1$ \textbf{to} $p$}
        \If {$order_s(sub^G_l) = |V^G|$ or $j = l$}  
            \State continue
        \EndIf
        \For{$h$ \textbf{from} $1$ \textbf{to} $order_s(sub^G_j)$}
            \State $v_{left} \leftarrow$ the $h$-th vertex in $V^G_j$
            \For{$f$ \textbf{from} $1$ \textbf{to} $order_s(sub^G_l)$}
                \State $v_{right} \leftarrow$ the $f$-th vertex in $V_l^G$
                \State $id \leftarrow |Sub^G_i|+|Sub_{n}|+1$
                \State $e^G_{id} \leftarrow$ build edge from $v^G_{left}$ to $v^G_{right}$
                \State $sub^G_{id} \leftarrow$ build novel subgraph by connecting $sub^G_j$ and $sub^G_l$ with $e^G_{id}$
                \If{$sub^G_{id}$ is generated before}
                    \State continue
                \EndIf
                \State $E_{n},Sub_{n}\leftarrow E_{n}\cup \{e^G_{id}\},Sub_{n}\cup \{sub^G_{id}\}$
            \EndFor
        \EndFor
    \EndFor
  \EndFor
  \State $Sub^G_{i+1}\leftarrow Sub^G_i \cup Sub_{n}$
  \State $E^{G_{i+1}}\leftarrow E^{G_i} \cup E_{n}$
  \State $G_{i+1}\leftarrow (V^G,E^{G_{i+1}})$
  \State return $Sub^G_{i+1}, G_{i+1}$
  \end{algorithmic}
\end{algorithm}

To model the generation of representation in a SAN-based model with $n$ layers, we split the entire generation process of MoG into $n$ steps. The $i$-th step of MoG generation is revealed to the generation of representations in the $i$-th SAN-based model layer. We define $G_i=(V^{G}, E^{G_i})$ as an intermediate state of MoG generated in the $i$-th step which is revealed to the representation generated in the $i$-th layer. Subgraphs generated in $(i-1)$-th step are used to generate new edges and subgraphs in $i$-th step, and $Sub^G_i=\{sub^G_1,...,sub^G_p\}$ is a set of subgraphs updated in the $i$-th step. Note that the $G_0$ has only subgraphs with one vertex and no edge which is revealed to the embeddings of SAN-based model. Algorithm \ref{alg:r1} demonstrates the procedure to generate the $G_{i+1}$ and  shows that the order of subgraphs generated in the $i$-th step is no more than $2^i$ which reflects the relationships between the ability to capture structural information and the layer. 

The condition to stop the generation is different for various SAN-based models. For example, the condition for the Transformer \cite{DBLP:conf/nips/VaswaniSPUJGKP17} with $n$ layers is that $G_n$ is finished, while the Universal Transformers\cite{DBLP:conf/iclr/DehghaniGVUK19} uses different numbers of steps for sentences with different lengths.

\begin{figure*}
\centering
\includegraphics[scale=0.25]{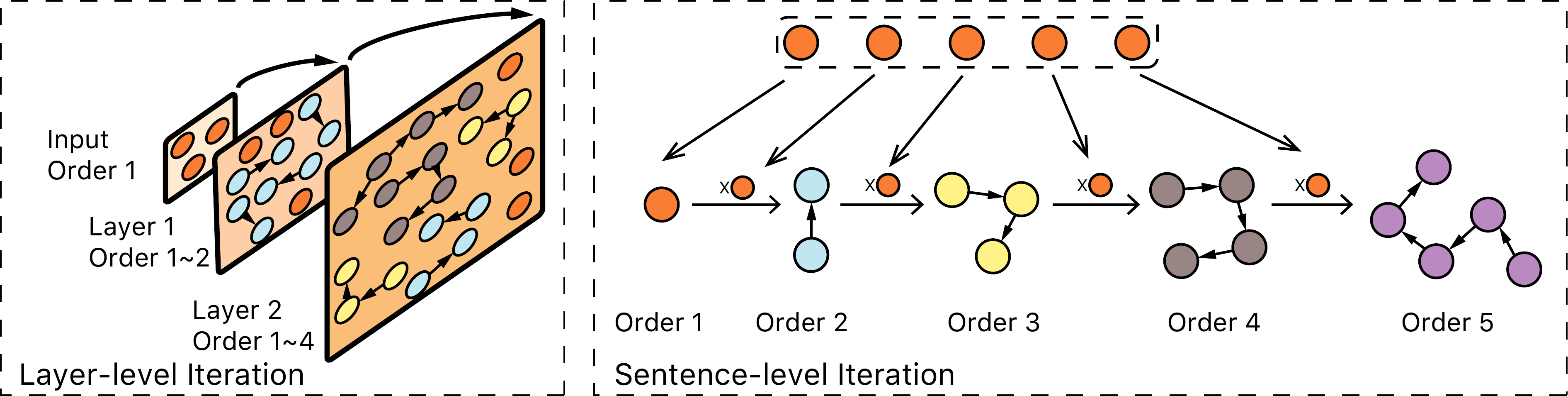}
\caption{Layer-level and sentence-level iteration. Layer-level iteration means that the model uses subgraphs generated in previous layers to build new subgraphs, and sentence-level iteration means that the model uses subgraphs generated in one layer to build new subgraphs in the same layer.}
\label{fig:problem2}
\end{figure*}

\subsection{MoG in Decoder}
Most seq2seq models are composed of encoders and decoders. The encoder and decoder with similar neural network structures can be viewed as two language models with similar structures to generate representations of source and target sentences, respectively. Same as encoder, the generation of representations in a decoder can also be viewed as a generation of MoG. However, MoG generated by the decoder is different from the encoder because it should reflect relationships between source and target sentences.

Given a source sentence $S=\{s_1,...,s_n\}$ and a target sentence $T=\{t_1,...,t_m\}$, $G^S=(V^S, E^S)$ is the MoG in the encoder, same as Section \ref{subsection:multigraph}, and we use $G^T=(V^T, E^T, G^S, E^{ST})$ to describe MoG in the decoder where $v^T_i \in V^T$ reflects the target word $t_i$, $e^{T}_j \in E^T$ connects $v^T_k$ and $v^T_l$, and $e^{ST}_j\in E^{ST}$ connects $v^T_k$ and $v^S_l$.  It is easy to know that subgraphs $sub^{G^T}_i \in Sub^{G^T}$ always consist of at least one $sub^{G^S}_j\in Sub^{G^S}$. Note that $e^{ST}_j$ is always from $v^S_k$ to $v^T_l$ and built after generation of $G^S$. If we use vertexes in $V^T$ and $V^S$ and edges in $E^{ST}$ to build a new graph, it is easy to know the new graph is bipartite.

The decoder is modified to focus on subsequent positions to reflect the direction of the sequence. Influenced by this ability, edges $e^T_i \in E^T$ are only from $v^T_j$ to $v^T_l$ with $j>l$. Out degrees and in degrees of $e^T_i \in E^T$ are based on word positions reflected by nodes. 

According to \cite{DBLP:conf/nips/SutskeverVL14}, the goal of the seq2seq model is to estimate the conditional probability 
\begin{equation}
    p(T=\{t_1,...,t_n\}|S=\{s_1,...,s_m\})
\end{equation}
where $n$ may differ from $m$. Model obtains representation of $S$ and computes the probability of $T$ as
\begin{equation}
    \begin{aligned}
    &p^{T'}_i=p(t_i|S,t_1,...,t_{i-1})\\
    &p(T'=\{t_1,...,t_n\}|S=\{s_1,...,s_m\})=\prod^{T'}_{i=1}p^{T'}_i
    \end{aligned}
    \label{eq:pros}
\end{equation}

As Eq (\ref{eq:pros}) shows, estimating conditional probability $p^{T'}_i$ can also be regarded as goal of training and decoding. Viewed representation as MoG, Eq (\ref{eq:pros}) can also equal to
\begin{equation}
    \begin{aligned}
    &p^{T'}_i=p(sub^{T'}_{t_i}|G^S,sub^{T'}_{t_1},...,sub^{T'}_{t_{i-1}})\\
    &p(G^{T'}|G^S)=\prod^{T'}_{i=1}p^{G^{T'}}_i
    \end{aligned}
    \label{eq:pros2}
\end{equation}
where $T'$ is the generated target sentence, $G^{T'}$ is the MoG generated by the decoder according to $T'$, and $sub^{T'}_{i}$ is the subgraph of $G^{T'}$. 

MoG shows that the seq2seq model implements Eq (\ref{eq:pros}) by incorporating graph structures among source sentences into structures of target sentences to capture relationships among subsequences of source and target.

\subsection{Two Questions for Various Models}
\label{section:questions}

MoG explanation can extend to RNN-based, CNN-based, and other models by viewing encoding as the generation of subgraphs with differences. Based on the generation of subgraphs, there are two fundamental questions for different models to classify.

$\bullet$ \textbf{How to implement iterative encoding?} Fig. \ref{fig:problem2} shows two kinds of iteration. Sentence-level iteration allows the model to encode words one by one, as  in the RNN-based model. With sentence-level iteration, the order of the subgraph is the sentence length. All layer-based models implement layer-level  iteration by generating representations in a layer and feeding them into the following layer.

$\bullet$ \textbf{How to capture edges and subgraphs?} RNN-based models use recurrent networks, CNN-based models \citep{DBLP:conf/icml/GehringAGYD17,DBLP:conf/icml/DauphinFAG17} use convoluation+gating blocks, and SAN-based models use self-attention.

The model performance may also be influenced by other factors, such as the dimensions or architecture of the model, and it is difficult to classify models by them.

\subsection{Multi-order-Graph in SAN-based Models}
\label{section:mog}

The SAN-based model is based on self-attention. The input of attention contains queries ($Q$), keys ($K$), and values ($V$) of input sequences. The attention is generated using queries and keys like Equation (\ref{orisof}),
\begin{equation}
\label{orisof}
\text{Attention}(Q, K, V) = {\rm softmax}(   
Q{K^\top}
/\sqrt{d_k})V.
\end{equation}
where $d_k$ is the dimension of $Q$, $K$, $V$.

SAN-based models use self-attention to capture edges and subgraphs, and use layer-level iteration only. Regarding representation as an MoG, we can use a vector for representation to contain all information in the MoG, which means that we may use a vector with the same shape as representation to represent a subgraph. 
Given a representation reflecting a set of subgraphs, a  representation pair can be presented by a set of subgraph pairs. Given representations $r_a^i$ and $r_b^i$ generated by the $i$-th layer, $\{sub_1^{a(i)},...,sub_n^{a(i)}\}$ are $n$ subgraphs to reflect $r_a^i$ and $\{sub_1^{b(i)},...,sub_m^{b(i)}\}$ are $m$ subgraphs to reflect $r_b^i$. Using ${\rm R}(sub_i)$ for the representation to represent subgraph $sub_i$, $r_a^i$ and $r_b^i$ can be represented as

\begin{equation}
r_a^i =\sum_j^n {\rm R}(sub_j^{a(i)}), r_b^i=\sum_j^m {\rm R}(sub_j^{b(i)}).
\end{equation}

Self-attention has to get an attention matrix $\mathcal{M}$ using queries and keys according to Equation (\ref{orisof}). In the $i$-th layer of the SAN-based model, given a sentence $S=\{s_1,...,s_n\}$, the attention matrix generated by self-attention is $\mathcal{M}_i$, representation of word $s_m$ generated by this layer is $r_m^i$. Attention $a_{kl}^i$ in the $k$-th row and $l$-th column of matrix $\mathcal{M}_i$ is calculated using $r_k^i$ as query and $r_l^i$ as key,
\begin{equation}
    \begin{aligned}
        a_{kl}^i&=
        r_l^i \cdot (r_k^i)^\top =(\sum_j^n {\rm R}(sub_j^{l(i)}) \cdot (\sum_p^m {\rm R}(sub_p^{k(i)}))^\top \\&=\sum_j^n \sum_p^m {\rm R}(sub_j^{l(i)}) \cdot ({\rm R}(sub_p^{k(i)}))^\top
    \end{aligned}
    \label{eq:submatrixs}
\end{equation}

Every ${\rm R}(sub_j^{l(i)}) \cdot ({\rm R}(sub_p^{k(i)}))^\top$ can be reflected by a edge in MoG.  Equation (\ref{eq:submatrixs}) shows that the attention score can be viewed as a sum of relationships between different parts of representation, which can be reflected by a group of edges in MoG. By putting all representation parts together as a representation, self-attention calculates all these relationships at once.

After generating all edges which reflect the attention score in $\mathcal{M}$, self-attention uses $M\cdot V$ to compute the representation, which can be viewed as the generation of subgraphs. Generated representation is a sum of different parts of representations, which can be viewed as a combination of subgraphs.

In the SAN-based model, a layer generating representation is a Generation Step of MoG in Algorithm \ref{alg:r1}, and the $i$-th layer corresponds to the $i$-th generation step. Feeding the representation of the $i$-th layer to the following layer is equal to feeding $G_{i+1}$ and $Sub_{i+1}^G$ to the $(i+1)$-th generation step.

In the $i$-th layer, representations used as query, key, and value are from the ($i$-1)-th layer, which means that subgraphs generated by the $(i-1)$-th layer will affect the highest order of subgraphs in the $i$-th layer. Connecting two input subgraphs of the highest order will generate a subgraph of the highest order in the  $i$-th layer, which makes the highest order of subgraphs increase exponentially as layers increase, and the highest order of subgraphs in the $i$-th layer is $2^i$.

However, it is quite likely that the SAN-based model cannot accurately capture all subgraphs because the highest order of subgraphs is limited by the number of layers. The outcome obtained by the $n$ -layer model may be incomplete if the input length exceeds $2^n$. Dehghani et al.\citep{DBLP:conf/iclr/DehghaniGVUK19} added a dynamic per-position halting mechanism to choose the required number of refinement steps, allowing the model to generate subgraphs of different order based on the input sentence.

\section{Graph-Transformer}
\begin{figure}
    \centering
    \includegraphics[scale=0.35]{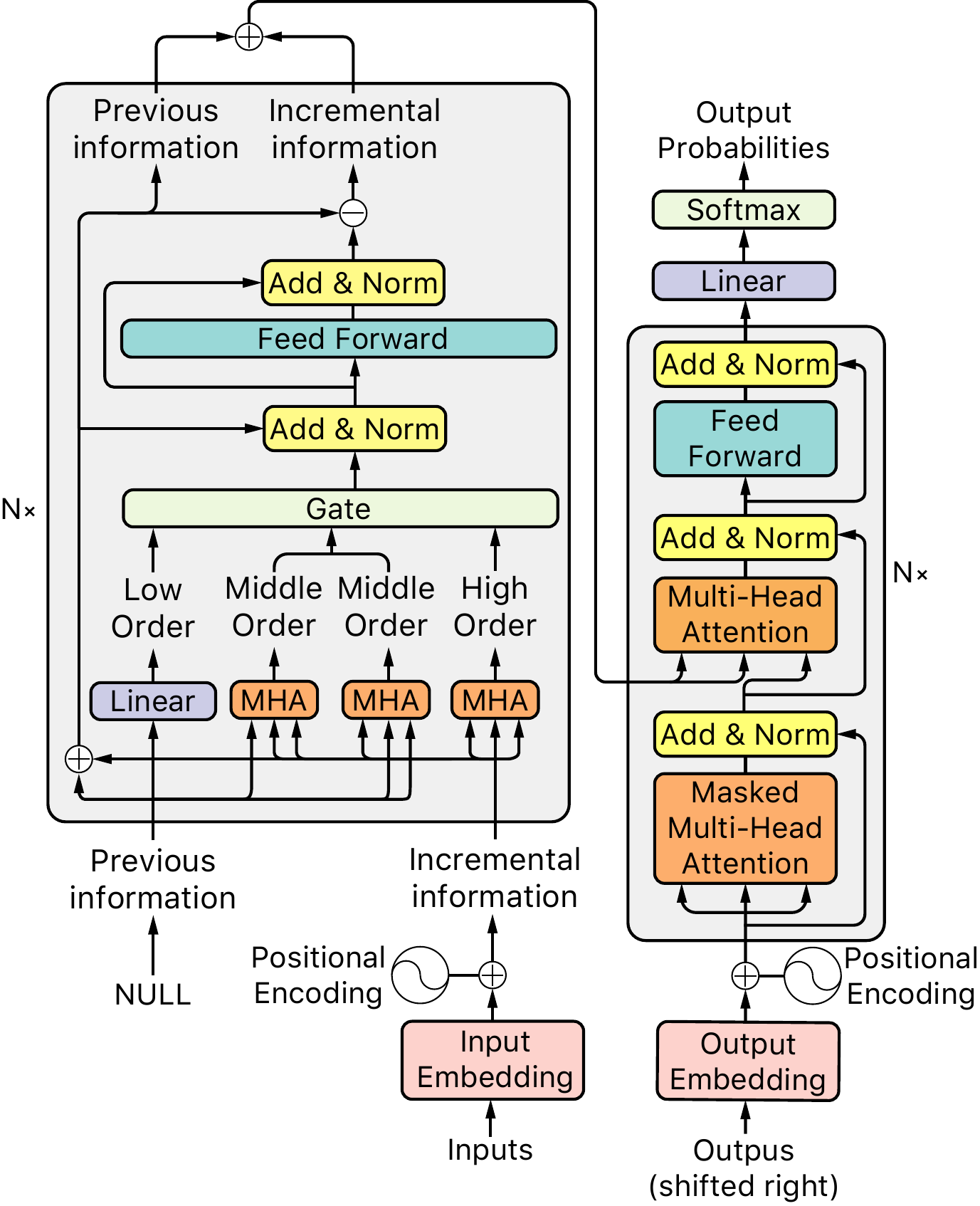}
    \caption{Graph-Transformer.}
    \label{fig:models_label}
\end{figure}

Modeling SAN-based encoder by MoG, models may capture subgraphs of low order repeatedly. A generated subgraph is always contained in representation and used to generate new subgraphs. As a result, the earlier subgraph is generated, the more times it will be generated. The weight of subgraphs of low-order increases in a latent way. Besides, saving multiple information of subgraphs in one vector makes it difficult for the model to distinguish them and hardly extract salient subgraphs from vectors. To solve this problem, we propose Graph-Transformer, where \textit{Graph} is revealed to subgraph of MoG and the input and output are still strings, to balance the weights of subgraphs and improve the performance of the SAN-based model.

\subsection{Self-Attention Group for Subgraphs of Different Orders} 
\label{section:41}
Fig. \ref{fig:models_label} illustrates the overall architecture of our Graph-Transformer. 

The original SAN-based model uses input representation as query, key, and value to calculate self-attention. To distinguish the representation learned in different steps, we first decompose the representations into different functional parts. We define the representation generated in a layer as \textbf{full representation}, and split the full representation into \textbf{previous representation} and \textbf{incremental representation}. In the $i$-th layer, the previous representation is the input of the full representation, reflecting the subgraphs generated before the $i$-th layer. Furthermore, in the $i$-th layer, the incremental representation is the difference between the full representation and the previous representation, reflecting new subgraphs generated in the $i$-th layer.

For the ($j$+1)-th layer, the calculation of self-attention can be viewed as the sum of four parts:
\begin{equation}
\begin{aligned}
r_f^j \cdot (r_f^j)^\top&= (r_p^j + r_i^j) \cdot (r_p^j+r_i^j)^\top \\
&=r_p^j \cdot (r_p^j)^\top + r_p^j \cdot (r_i^j)^\top + r_i^j \cdot (r_p^j)^\top+r_i^j \cdot (r_i^j)^\top
\end{aligned}
\end{equation}
\noindent where $r^j_f$, $r^j_p$, and $r^j_i$ are the full representation, previous representation, and incremental representation of the $j$-th layer. In the SAN-based model, $r_f^j$ is the vector generated in the $j$-th layer, $r_p^j$ is the vector generated in the $(j-1)$-th layer or the input vector in the $j$-th layer, and $r_i^j$ is equal to $r_f^j-r_p^j$.

Note that $r_p$ is also $r_f$ of the previous layer, which means that $r_p \cdot r_p^\top$ has been calculated by the previous layer and makes subgraphs be generated repeatedly. It is also the key to increasing the weight of low-order subgraphs. To avoid redundancy, we only calculate the other three parts of self-attention. There are three levels for the subgraph order:

$\bullet$ \textbf{High order.} Subgraphs generated by $r_i \cdot r_i^\top$ belong to high order, and a part of self-attention is used to process high-order subgraphs, which uses input incremental representation as to its query, key, and value. In the $i$-th layer, the order of subgraphs is in the range of $2^{i-1}$ to $2^i$.

$\bullet$ \textbf{Middle order.} Subgraphs generated by $r_p \cdot r_i^\top$ and $r_i \cdot r_p^\top$ belong to the middle order and the other two parts of self-attention. The second part of self-attention uses incremental representation as query and previous representation as key and value. The third part of self-attention uses previous representation as query and incremental representation as key and value. In the $i$-th layer, the order of subgraphs is in the range of $2^{i-2}$ to $2^{i-1}$. 

$\bullet$ \textbf{Low order.} Subgraphs generated by $r_p \cdot r_p^\top$ belong to low order. As discussed above, it is no need to calculate $r_p \cdot r_p^\top$ again. Instead of self-attention, we use a linear function to transform vector space. The subgraph order is in the range of 1 to $2^{i-2}$. We calculate representation as
\begin{equation}
    \begin{aligned}
    h_{j+1}^{h}=&\text{Attention}^{h}_j(Q^i_j,K^i_j,V^i_j),\\
    h_{j+1}^{m}=&\text{Attention}^{m}_j(Q^i_j,K^p_j,V^p_j)+\\ &\text{Attention}^{m}_j(Q^p_j,K^i_j,V^i_j),\\
    h_{j+1}^{l}=&\text{Linear}(r^p_j),\\
    h_{j+1}^{f}=&\text{LayerNorm}(r_j^f+\text{fusion}(h_{j}^h,h_{j+1}^m,h_{j+1}^l)),\\
    r_{j+1}^f=&\text{LayerNorm}(h_{j+1}^f+\text{FFN}(h^f_{j+1})),\\
    r_{j+1}^i=&r_{j}^f-r^f_j, r_{j+1}^p=r^f_j\\
    \end{aligned}
\end{equation}

\noindent where $Q^i_j$, $Q^p_j$ and $Q^f_j$ are query vectors for $r_j^i$, $r_j^p$ and $r_f^p$ respectively, and same superscript and subscript for key vectors $K$ and value vectors $V$, and $h^h_{j+1}$, $h^m_{j+1}$, and $h^l_{j+1}$ are the hidden states for high-order, middle-order, and low-order subgraphs, respectively.

To reduce the number of parameters and avoid overfitting, we use method \textbf{shared-query-key-value} to share query, key, and value vectors in different parts of self-attention, while it is difficult to train such a model because different groups of subgraphs require different vector spaces. To keep the least effect on the performance, we also use method \textbf{half-dimension} to reduce the dimension of model to half of the original dimension.

\subsection{Fusion of Representations}
To get the full representation, we introduce three fusing strategies to combine previous and incremental representations. 

\textbf{Sum}. Calculating the sum is the most straightforward  strategy. However, this strategy depends on the quality of previous and incremental representations. Besides, the model gives four groups of subgraphs equal weights, which cannot indicate the important subgraphs.

\textbf{Weight-gate}. Representations generated by self-attention are new subgraphs that have not been weighted by the model. Viewing these three parts of representation as a group, we can use a gate to calculate their importance and merge them.

\begin{equation}
\begin{aligned}
w=&{\rm Sigmoid}(h_h+h_m+h_l), r_f
\\
=&(h_h+h_m) \cdot w + h_l\cdot(1-w),
\end{aligned}
\end{equation}

\noindent Using a gate to assign weights, the model can explicitly distinguish new and old subgraphs and pay attention to important groups of subgraphs. The disadvantage of this method is that the model cannot distinguish subgraphs of high and middle orders. We call this strategy \textbf{weight-gate}.

\textbf{Self-gate}. Wang et al.\cite{wang-etal-2018-multi-layer} proposed a fusion function based on self-attention with hops for fusing representations from different layers.  Similarly, we use self-attention to generate a matrix of weight which stands for relationships among representations. To assign the weight of three parts of representations, we concatenate four representations to form a new sequence $R$ and calculate the relationship matrix.

\begin{equation}
\begin{aligned}
R&=\text{concat}(\left[h_{j}^h,h_{j}^m,h_{j}^l\right]),\\
R_q&=RW^Q,R_k=RW^K,R_v=RW^V,\\
R_f&=\text{softmax}(R_q{R_k^T}/d_k) R_v/ 4,
\end{aligned}
\end{equation}
where $R_f$ is the representation sequence, $R_q$, $R_k$ and $R_v$ are query, key, and value vectors, and $d_k$ is the model dimension. This method can capture relationship between representations and weight them. The weight of a group will gain if it is more important than others. To make the sum of weights equal to 1, the representation is divided by 4. We call this strategy \textbf{self-gate}. Self-gate can weight all representations by the model, while according to the property of self-attention, self-gate will generate higher-order subgraphs, which makes the model deeper and more difficult to train.

\begin{table*}[htb]

\newcommand{\tabincell}[2]{\begin{tabular}{@{}#1@{}}#2\end{tabular}}
\caption{Multi-BLEU scores on four NMT tasks. Note that FF is short for the dimension of feed-forward. Results with $^\ast$ present statistically significant differences (p $<$ 0.05). } 

\begin{center}
\begin{tabular}{l|l|l|l|l}
\toprule
\multirow{2}{*}{\vspace{-2mm}Model} &    \multicolumn{1}{c|}{De-En}&\multicolumn{1}{c|}{En-De} &\multicolumn{1}{c|}{En-Fr}&\multicolumn{1}{c}{En-Ro}\\
\cmidrule{2-5}

&  BLEU &   BLEU &  BLEU &  BLEU  \\
\hline
\multicolumn{5}{c}{Existing NMT systems}\\
\hline
Transformer (base)\cite{shaw-etal-2018-self}&-&26.5&38.2&-\\
Relative Position Encoding\cite{shaw-etal-2018-self}&-&26.8 (+0.3)&38.7 (+0.5)&-\\
\hdashline
Transformer (small) \citet{DBLP:conf/nips/HeTXHQ0L18}&32.9&-&-&-\\
Transformer (base) \citet{DBLP:conf/nips/HeTXHQ0L18}&32.9&27.3&-&-\\

Layer-wise Coordination\cite{DBLP:conf/nips/HeTXHQ0L18}&35.1 (+2.2)&28.3 (+1.0)&-&-\\
Reformer\cite{DBLP:conf/iclr/KitaevKL20}&-&28.0 (+0.7)&-&-\\

\hline
\multicolumn{5}{c}{Our NMT systems}\\
\hline
Transformer(base, FF-1024)   &  36.5 & - &- &-\\
Transformer(base) &  -& 27.1 &43.3 & 33.9 \\

\hline
\tabincell{c}{Graph-Transformer } & \textbf{37.6$^\ast$ (+1.1)} & \textbf{28.3$^\ast$ (+1.2)}& \textbf{44.8$^\ast$ (+1.5)}&\textbf{35.0$^\ast$ (+1.1)}\\
\hline
\end{tabular}
\end{center}

\label{bleu1}
\end{table*}

\begin{table*}[htb]

\newcommand{\tabincell}[2]{\begin{tabular}{@{}#1@{}}#2\end{tabular}}
\caption{Multi-BLEU scores of ablations on De-En and En-De. \#Para, \#Speed, \#Mem and \#PPL denote the size of model paragraphs, training speed (tokens/second), GPU memory model used (GB) and perplexity respectively.}

\begin{center}
\begin{tabular}{l|ccc|ccc}
\toprule
\multirow{2}{*}{\vspace{-2mm}Model} &    \multicolumn{3}{c|}{De-En}&\multicolumn{3}{c}{En-De}\\
\cmidrule{2-7}

&  BLEU & \#Para &\#Speed& BLEU & \#Para  &\#Speed \\
\hline
Transformer(base, FF-1024)   &  36.5 & 42M &50K&- &- &-\\
Transformer(base) &  -&-&-& 27.1 &66M& 137K\\

\hline
\tabincell{c}{sum} & 37.1 &50M&42K& 27.5 &77M&112K\\
\tabincell{c}{weight-gate} & 37.3&57M &39K &  28.0 & 80M&109K\\
\tabincell{c}{self-gate} &36.9&53M&30K& 27.6 & 77M&91K\\
\hdashline
 weight-gate\&shared-qkv  &37.1&51M&40K&27.7&75M&121K\\
 weight-gate\&half-dim     & 37.6 & 50M&35K& 28.3 & 74M&111K\\
 weight-gate\&half-dim     \&shared-qkv  &37.5&47M&38K&27.7&70M&115K\\
\hline

\end{tabular}
\end{center}
 
\label{bleu1ablation}

\end{table*}

\section{Experiments and Results}
\label{results}
Our graph-Transformer will be mainly evaluated on four NMT tasks, IWSLT14 German-English (De-En), WMT14  English-German (En-De), WMT14 English-French (En-Fr), and WMT16 English-Romanian (En-Ro).

\subsection{Datasets}

\noindent\textbf{IWSLT14 De-En } IWSLT14 De-En dataset contains 153K training sentence pairs. We use script\footnote{https://github.com/pytorch/fairseq/blob/master/examples/\\translation/prepare-iwslt14.sh} to preprocess the dataset, and use 7K data from the training set as the validation set and the combination of dev2010, dev2012, tst2010, tst2011 and tst2012 as the test set with 7K sentences. BPE algorithm is used to process words into subwords, and the number of subword tokens is 10K.

\noindent\textbf{WMT14 En-De,  En-Fr and WMT16 En-Ro} WMT14 En-De, WMT14 En-Fr, and WMT16 En-Ro datasets with 4.5M, 36M, and 610K sentence pairs are used for training. For En-De and En-Fr, we use 7K and 26K data from the training set as the validation set respectively, and newstest2014 as the test set. We use script\footnote{https://github.com/pytorch/fairseq/blob/master/examples/\\translation/prepare-wmt14en2de.sh} and script \footnote{https://github.com/pytorch/fairseq/blob/master/examples/\\translation/prepare-wmt14en2fr.sh} for En-De and  En-Fr,  respectively. For En-Ro, we use the test2013 for validation, and test2016 as the test set. The  sentences longer than 250 are removed from the training dataset. Dataset is segmented by BPE so that the shared vocabulary has 40K  subwords.

\subsection{Model Configurations}
For De-En, our model uses 6 encoder and decoder layers with the model dimension of 512, the feed-forward dimension of 1024 and dropout of 0.3. For En-De, En-Fr, and En-Ro, our model uses 6 encoder and decoder layers with the model dimension of 512, the feed-forward dimension of 2048 and dropout of 0.1.
\subsection{Training of Experiment}

Our models for En-De, En-Fr and En-Ro are trained on one CPU (Intel i7-5960X) and 
four nVidia RTX TITAN X GPUs
, and our models for De-En are trained on the same CPU and one nVidia RTX TITAN X GPU. The model implementation for NMT tasks is based on fairseq-0.6.2\footnote{https://github.com/facebookresearch/fairseq}. We choose Adam optimizer with $\beta_1=0.9$, $\beta_2=0.98$, $\epsilon=10^{-9}$ and the learning rate setting strategy, which are all the same as \cite{DBLP:conf/nips/VaswaniSPUJGKP17}.

\begin{equation}
lr = d^{-0.5} \cdot \text{min}(step^{-0.5}, step \cdot warmup_{step}^{-1.5})
\end{equation}

\noindent where $d$ is the dimension of embeddings, $step$ is the step number of training,  and $warmup_{step}$ is the step number of warmup. When the step number of training is smaller than the step number of warmup, the learning rate increases linearly and then decreases. We set $warmup_{step}$ as 4000 for En-De and De-En, and 8000  for En-Fr and En-Ro.

The batch size is 1024 for De-En and 4096 for En-De, En-Fr and En-Ro. We use the beam search decoder for De-En  with beam width $5$. For En-De, En-Fr, and En-Ro, following \cite{DBLP:conf/nips/VaswaniSPUJGKP17}, the  beam width  is $4$, and the length penalty $\alpha$ is 0.6. We evaluate the translation results by using tokenized BLEU \cite{DBLP:conf/acl/PapineniRWZ02} score calculated with the \texttt{multi-bleu.perl} script. Statistical significance ($p < 0.05$) on the difference of
BLEU scores is tested by \textit{bootstrap-hypothesis-difference-significance.pl}\footnote{https://github.com/moses-smt/mosesdecoder/blob/master/scripts/analysis/\\bootstrap-hypothesis-difference-significance.pl}.

\subsection{Results}

\begin{figure}[htp]

\setlength{\abovecaptionskip}{0pt}
\centering
\pgfplotsset{height=5.5cm,width=10.5cm,compat=1.15,every axis/.append style={thick},every axis legend/.append style={at={(0.67,1)}},legend columns=2}
\subfigure[]{
\begin{tikzpicture}
\tikzset{every node}
\begin{axis}
[height=4.5cm,width=5cm,enlargelimits=0.08, tick align=inside, 
 xticklabels={ $10$,$20$,$30$,$40$,$50$,$50+$},
xtick={0,1,2,3,4,5},
ylabel={BLEU},xlabel={Length of Sentence},
legend to name=legend,]
\addplot+ [sharp plot, mark=o,mark size=1.5pt,mark options={mark color=red}, color=red] coordinates
{(0,25.7)(1,26.9)(2,26.9)(3,27.1)(4,27.8)(5,29)};
\addlegendentry{baseline}
\addplot+ [sharp plot, mark=square*,mark size=1.5pt,mark options={mark color=blue}, color=blue] coordinates
{(0,26.3)(1,27.5)(2,28.3)(3,27.7)(4,29.0)(5,31.0)};
\addlegendentry{gate+half-dim}
\end{axis}
\label{fig:length}
\end{tikzpicture}
}\subfigure[]{
\begin{tikzpicture}
\tikzset{every node}
\begin{axis}
[height=4.5cm,width=5cm,enlargelimits=0.08, tick align=inside, 
 xticklabels={ $3$,$4$,$5$,$6$,$7$},
xtick={1,2,3,4,5},
xlabel={Number of Layers}]
\addplot+ [sharp plot, mark=o,mark size=1.5pt,mark options={mark color=red}, color=red] coordinates
{(1,26.3)(2,26.675)(3,27.15)(4,27.1)(5,27.7)};
\addplot+ [sharp plot, mark=square*,mark size=1.5pt,mark options={mark color=blue}, color=blue] coordinates
{(1,26.7)(2,27.1)(3,27.8)(4,28.3)(5,28.5)};
\end{axis}
\label{fig:layer}
\end{tikzpicture}
}
\begin{center}
   \ref{legend} 
\end{center}
\caption{BLEU points of different lengths and models with different numbers of layers.}
\label{fig:bleulengths}
\end{figure}

The baselines for En-De, En-Fr, and En-Ro are Transformer-base, and the  baseline for De-En is Transformer-base with the feed-forward dimension of 1024. Table \ref{bleu1}  compares our Graph-Transformer with the baseline, showing that our model enhances all tasks and outperforms all baselines. For De-En tasks, our model with half-dimension and weight-gate gets the best performance of 37.6 BLEU points outperforming the baseline by 1.1 BLEU points with 50 million parameters. For En-De tasks, our model with half-dimension and weight-gate gets the best performance of 28.3 BLEU points outperforming the baseline by 1.2 BLEU points with 74 million parameters. 
For En-Fr and En-Ro, our model with half-dimension and weight-gate gets the performance of 44.8 and 35.0 BLEU points, outperforming the baseline by 1.5  and 1.1 BLEU points, respectively.  With a baseline of 27.1 BLEU points  on En-De and 43.3 BLEU points  on En-Fr, the improvement of Graph-Transformer is better than \citet{shaw-etal-2018-self} and \citet{DBLP:conf/nips/HeTXHQ0L18} on En-De and En-Fr tasks.

\begin{figure*}[htp]

\setlength{\abovecaptionskip}{0pt}
\flushleft
\begin{tikzpicture}
\tikzset{every node}
\begin{axis}
[width=5cm,enlargelimits=0.13, tick align=inside, 
xticklabels={ $2$,$3$,$4$,$5$,$6$,$7$,$8$,$9$,$10$,$11$,$12$},
xtick={0,1,2,3,4,5,6,7,8,9,10},
ylabel={Weight},xlabel={Layers},
legend columns=-1,
legend entries={0-10,10-20,20-30,30-40,40-50,50+},
legend to name=legend,
]
\addplot+ [sharp plot,mark=o,mark size=1.5pt,mark options={mark color=red}, color=red] coordinates
{(0,0.3815886705833333)(1,0.4141888795833333)(2,0.42452337533333334)(3,0.5090552310833335)};
\addplot+ [sharp plot, mark=square*,mark size=1.5pt,mark options={mark color=blue}, color=blue] coordinates
{(0,0.4245237229282406)(1,0.47115694674768455)(2,0.4452153016087967)(3,0.5518155045023154)};
\addplot+ [sharp plot, mark=diamond*,mark size=1.5pt,mark options={mark color=cyan}, color=cyan] coordinates
{(0,0.4449529435775864)(1,0.49047496687500003)(2,0.4549015632974142)(3,0.5693499775323276)};
\addplot+ [sharp plot, mark=pentagon*,mark size=1.5pt,mark options={mark color=pink}, color=pink] coordinates
{(0,0.45100725390284746)(1,0.5019389085427142)(2,0.45814783376884455)(3,0.5776272000335015)};
\addplot+ [sharp plot, mark=triangle*,mark size=1.5pt,mark options={mark color=violet}, color=violet] coordinates
{(0,0.45577986505263146)(1,0.5092601193333334)(2,0.4583225932280703)(3,0.5823491752982454)};
\addplot+ [sharp plot, mark=asterisk,mark size=1.5pt,mark options={mark color=brown}, color=brown] coordinates
{(0,0.4450976280382775)(1,0.5082943183253582)(2,0.4522527580861244)(3,0.5845326474641144)};
\end{axis}
\end{tikzpicture}
\begin{tikzpicture}[
line1/.style={sharp plot,mark=o,mark size=1.5pt,mark options={mark color=red}, color=red},
line2/.style={sharp plot, mark=square*,mark size=1.5pt,mark options={mark color=blue}, color=blue},
line3/.style={sharp plot, mark=diamond*,mark size=1.5pt,mark options={mark color=cyan}, color=cyan},
line4/.style={sharp plot, mark=pentagon*,mark size=1.5pt,mark options={mark color=pink}, color=pink},
line5/.style={sharp plot, mark=triangle*,mark size=1.5pt,mark options={mark color=violet}, color=violet},
line7/.style={sharp plot, mark=asterisk,mark size=1.5pt,mark options={mark color=brown}, color=brown}
]
\tikzset{every node}
\begin{axis}
[width=5cm,enlargelimits=0.13, tick align=inside, 
xticklabels={ $2$,$3$,$4$,$5$,$6$,$7$,$8$,$9$,$10$,$11$,$12$},
xtick={0,1,2,3,4,5,6,7,8,9,10},
xlabel={Layers}]
\addplot+ [line1] coordinates
{(0,0.34560567774999995)(1,0.40453156824999986)(2,0.41648590920833345)(3,0.4762465010833337)(4,0.4904235333333333)};
\addplot+ [line2] coordinates
{(0,0.3908409294907408)(1,0.4691752048958335)(2,0.44646039417245414)(3,0.5144427339988427)(4,0.5218100732175928)};
\addplot+ [line3] coordinates
{(0,0.40819531830818917)(1,0.4932445860183188)(2,0.4587682716594826)(3,0.5320431607273705)(4,0.5336287176831896)};
\addplot+ [line4] coordinates
{(0,0.41353863122278084)(1,0.5045647348911221)(2,0.46533600624790666)(3,0.5379218278643216)(4,0.5413390113902841)};
\addplot+ [line5] coordinates
{(0,0.4184684809999999)(1,0.5124243272280704)(2,0.46768552989473694)(3,0.5434356157368418)(4,0.5475895382807019)};
\addplot+ [line7] coordinates
{(0,0.40782839901913853)(1,0.5110494781578943)(2,0.4601179508851671)(3,0.5397754163397123)(4,0.5426716092822966)};
\end{axis}
\end{tikzpicture}
\begin{tikzpicture}[
line1/.style={sharp plot,mark=o,mark size=1.5pt,mark options={mark color=red}, color=red},
line2/.style={sharp plot, mark=square*,mark size=1.5pt,mark options={mark color=blue}, color=blue},
line3/.style={sharp plot, mark=diamond*,mark size=1.5pt,mark options={mark color=cyan}, color=cyan},
line4/.style={sharp plot, mark=pentagon*,mark size=1.5pt,mark options={mark color=pink}, color=pink},
line5/.style={sharp plot, mark=triangle*,mark size=1.5pt,mark options={mark color=violet}, color=violet},
line7/.style={sharp plot, mark=asterisk,mark size=1.5pt,mark options={mark color=brown}, color=brown}
]
\tikzset{every node}
\begin{axis}
[width=5cm,enlargelimits=0.13, tick align=inside, 
xticklabels={ $2$,$3$,$4$,$5$,$6$,$7$,$8$,$9$,$10$,$11$,$12$},
xtick={0,1,2,3,4,5,6,7,8,9,10},
xlabel={Layers}]
\addplot+ [line1] coordinates
{(0,0.2594607900833334)(1,0.4206781385833334)(2,0.3622195445833334)(3,0.44009341875000013)(4,0.40453454958333346)(5,0.43563733799999976)};
\addplot+ [line2] coordinates
{(0,0.29693320438657417)(1,0.4707481456828705)(2,0.4028281296875001)(3,0.500817962685185)(4,0.4364224473032403)(5,0.4633796284837959)};
\addplot+ [line3] coordinates
{(0,0.30809290507543113)(1,0.4934075673060341)(2,0.4226193515732758)(3,0.5201898376508624)(4,0.44742054554956917)(5,0.47776630990301744)};
\addplot+ [line4] coordinates
{(0,0.3112789887437186)(1,0.5006423057286434)(2,0.4325603079899504)(3,0.5265604388777217)(4,0.4505386526968171)(5,0.4888683094974874)};
\addplot+ [line5] coordinates
{(0,0.31518565442105273)(1,0.5065211155789475)(2,0.43961045017543876)(3,0.5288544863859646)(4,0.4544822698245613)(5,0.4926478559298247)};
\addplot+ [line7] coordinates
{(0,0.3068454945454546)(1,0.5013901992822969)(2,0.43367916339712886)(3,0.5218175980382774)(4,0.44806596555023914)(5,0.4938512797129184)};
\end{axis}
\end{tikzpicture}
\begin{tikzpicture}[
line1/.style={sharp plot,mark=o,mark size=1.5pt,mark options={mark color=red}, color=red},
line2/.style={sharp plot, mark=square*,mark size=1.5pt,mark options={mark color=blue}, color=blue},
line3/.style={sharp plot, mark=diamond*,mark size=1.5pt,mark options={mark color=cyan}, color=cyan},
line4/.style={sharp plot, mark=pentagon*,mark size=1.5pt,mark options={mark color=pink}, color=pink},
line5/.style={sharp plot, mark=triangle*,mark size=1.5pt,mark options={mark color=violet}, color=violet},
line7/.style={sharp plot, mark=asterisk,mark size=1.5pt,mark options={mark color=brown}, color=brown}
]
\tikzset{every node}
\begin{axis}
[width=5cm,enlargelimits=0.13, tick align=inside, 
xticklabels={ $2$,$3$,$4$,$5$,$6$,$7$,$8$,$9$,$10$,$11$,$12$},
xtick={0,1,2,3,4,5,6,7,8,9,10},
xlabel={Layers}]
\addplot+ [line1] coordinates
{(0,0.2979985462500002)(1,0.28523363399999996)(2,0.23125742458333334)(3,0.3695713672500001)(4,0.3537232710833333)(5,0.41779427633333327)(6,0.45073111783333364)};
\addplot+ [line2] coordinates
{(0,0.3116788058449072)(1,0.3447517189236115)(2,0.2926929506712963)(3,0.40191184805555524)(4,0.38521882773148175)(5,0.4609277831828706)(6,0.4713815806944443)};
\addplot+ [line3] coordinates
{(0,0.30579080377155166)(1,0.3662546181896556)(2,0.3201601249137935)(3,0.4181796626939657)(4,0.395920673092672)(5,0.47138335102370715)(6,0.4826689367672418)};
\addplot+ [line4] coordinates
{(0,0.2997662996314905)(1,0.37327014134003406)(2,0.3313609194472363)(3,0.42739557160804076)(4,0.4003675006197654)(5,0.4708605831825797)(6,0.4908425577051929)};
\addplot+ [line5] coordinates
{(0,0.2971796960701753)(1,0.3785380064912281)(2,0.3391548865614036)(3,0.43473998329824526)(4,0.4036701532631581)(5,0.475457144736842)(6,0.49634699207017546)};
\addplot+ [line7] coordinates
{(0,0.2888287533492824)(1,0.37467140660287085)(2,0.32991963645933037)(3,0.43542683186602843)(4,0.3932899220095693)(5,0.4675720088516748)(6,0.49306697531100496)};
\end{axis}
\end{tikzpicture}
\\
\begin{tikzpicture}[
line1/.style={sharp plot,mark=o,mark size=1.5pt,mark options={mark color=red}, color=red},
line2/.style={sharp plot, mark=square*,mark size=1.5pt,mark options={mark color=blue}, color=blue},
line3/.style={sharp plot, mark=diamond*,mark size=1.5pt,mark options={mark color=cyan}, color=cyan},
line4/.style={sharp plot, mark=pentagon*,mark size=1.5pt,mark options={mark color=pink}, color=pink},
line5/.style={sharp plot, mark=triangle*,mark size=1.5pt,mark options={mark color=violet}, color=violet},
line7/.style={sharp plot, mark=asterisk,mark size=1.5pt,mark options={mark color=brown}, color=brown}
]
\tikzset{every node}
\begin{axis}
[width=5cm,enlargelimits=0.13, tick align=inside, 
xticklabels={ $2$,$3$,$4$,$5$,$6$,$7$,$8$,$9$,$10$,$11$,$12$},
xtick={0,1,2,3,4,5,6,7,8,9,10},
ylabel={Weight},xlabel={Layers}]
\addplot+ [line1] coordinates
{(0,0.01171875)(1,0.31769147650000007)(2,0.2919477624166667)(3,0.28810157208333337)(4,0.31720957158333335)(5,0.40891100174999984)(6,0.3917784951666668)(7,0.4275116037500001)};
\addplot+ [line2] coordinates
{(0,0.01171875)(1,0.3529145418518518)(2,0.31663639915509295)(3,0.3437367664814814)(4,0.3521510600115741)(5,0.44065780137731536)(6,0.4146756555208335)(7,0.4707425818518524)};
\addplot+ [line3] coordinates
{(0,0.01171875)(1,0.36889322234913824)(2,0.3260668171228444)(3,0.3673539162176721)(4,0.36447912165948254)(5,0.45395121521551757)(6,0.4248766201077588)(7,0.4874212839224139)};
\addplot+ [line4] coordinates
{(0,0.01171875)(1,0.37812429021775573)(2,0.33364265318257985)(3,0.37909301544388585)(4,0.36760876693467337)(5,0.45734919683417047)(6,0.4299931465829148)(7,0.49639545663316603)};
\addplot+ [line5] coordinates
{(0,0.01171875)(1,0.3846688731929825)(2,0.3392789695438596)(3,0.38737884024561414)(4,0.36659399171929846)(5,0.46168895943859656)(6,0.4328593562807017)(7,0.5018096709473681)};
\addplot+ [line7] coordinates
{(0,0.01171875)(1,0.38350422698564607)(2,0.3375587680861244)(3,0.38196316277511944)(4,0.35554314937799053)(5,0.45329130602870776)(6,0.42751357138755974)(7,0.5030928764593299)};
\end{axis}
\end{tikzpicture}
\begin{tikzpicture}[
line1/.style={sharp plot,mark=o,mark size=1.5pt,mark options={mark color=red}, color=red},
line2/.style={sharp plot, mark=square*,mark size=1.5pt,mark options={mark color=blue}, color=blue},
line3/.style={sharp plot, mark=diamond*,mark size=1.5pt,mark options={mark color=cyan}, color=cyan},
line4/.style={sharp plot, mark=pentagon*,mark size=1.5pt,mark options={mark color=pink}, color=pink},
line5/.style={sharp plot, mark=triangle*,mark size=1.5pt,mark options={mark color=violet}, color=violet},
line7/.style={sharp plot, mark=asterisk,mark size=1.5pt,mark options={mark color=brown}, color=brown}
]
\tikzset{every node}
\begin{axis}
[width=5cm,enlargelimits=0.13, tick align=inside, 
xticklabels={ $2$,$ $,$4$,$ $,$6$,$ $,$8$,$ $,$10$,$ $,$12$},
xtick={0,1,2,3,4,5,6,7,8,9,10},
xlabel={Layers}]
\addplot+ [line1] coordinates
{(0,0.00390625)(1,0.249051616)(2,0.26136290625)(3,0.3067685763333333)(4,0.2937462261666666)(5,0.3023334473333333)(6,0.35695817108333333)(7,0.36679723524999985)(8,0.4130507353333333)};
\addplot+ [line2] coordinates
{(0,0.00390625)(1,0.2846512982754631)(2,0.30224342457175957)(3,0.34618559393518555)(4,0.32550579293981496)(5,0.34887537748842595)(6,0.38332319979166674)(7,0.4039646696759262)(8,0.4439647738194448)};
\addplot+ [line3] coordinates
{(0,0.00390625)(1,0.29761726985991416)(2,0.31963431012931054)(3,0.36681084721982765)(4,0.3356180794396551)(5,0.3696471643426727)(6,0.39070874707974174)(7,0.4187163189331896)(8,0.4587613228232757)};
\addplot+ [line4] coordinates
{(0,0.00390625)(1,0.3043726827135678)(2,0.32914964634840843)(3,0.37361277170854235)(4,0.34101695257956444)(5,0.3773701182914574)(6,0.39113274011725246)(7,0.4253209100837523)(8,0.4689789425628142)};
\addplot+ [line5] coordinates
{(0,0.00390625)(1,0.30504754364912273)(2,0.33643015238596513)(3,0.377850882)(4,0.34449167259649116)(5,0.3805436444561406)(6,0.3929284700350877)(7,0.4317281234736845)(8,0.4759304805614034)};
\addplot+ [line7] coordinates
{(0,0.00390625)(1,0.2992927897129188)(2,0.33156873186602864)(3,0.37197085846889966)(4,0.3368900260765549)(5,0.37328973569377966)(6,0.3822322674641151)(7,0.43097115315789447)(8,0.4723212603827754)};
\end{axis}
\end{tikzpicture}
\begin{tikzpicture}[
line1/.style={sharp plot,mark=o,mark size=1.5pt,mark options={mark color=red}, color=red},
line2/.style={sharp plot, mark=square*,mark size=1.5pt,mark options={mark color=blue}, color=blue},
line3/.style={sharp plot, mark=diamond*,mark size=1.5pt,mark options={mark color=cyan}, color=cyan},
line4/.style={sharp plot, mark=pentagon*,mark size=1.5pt,mark options={mark color=pink}, color=pink},
line5/.style={sharp plot, mark=triangle*,mark size=1.5pt,mark options={mark color=violet}, color=violet},
line7/.style={sharp plot, mark=asterisk,mark size=1.5pt,mark options={mark color=brown}, color=brown}
]
\tikzset{every node}
\begin{axis}
[width=5cm,enlargelimits=0.13, tick align=inside, 
xticklabels={ $2$,$ $,$4$,$ $,$6$,$ $,$8$,$ $,$10$,$ $,$ $},
xtick={0,1,2,3,4,5,6,7,8,9,10},
xlabel={Layers}]
\addplot+ [line1] coordinates
{(0,0.00390625)(1,0.22938939633333325)(2,0.036883422166666666)(3,0.2560406712499999)(4,0.2430480865833333)(5,0.25194524500000004)(6,0.2523705819166667)(7,0.33916050816666665)(8,0.3861141174166667)(9,0.4199366616666669)};
\addplot+ [line2] coordinates
{(0,0.00390625)(1,0.24443771774305578)(2,0.04076320042824076)(3,0.282052225972222)(4,0.2809815663773144)(5,0.28762107427083355)(6,0.29274469798611114)(7,0.37018774189814846)(8,0.41259478967592605)(9,0.4734274995601849)};
\addplot+ [line3] coordinates
{(0,0.003906250452586207)(1,0.25152642968749983)(2,0.042464031443965496)(3,0.2917594669504312)(4,0.29527431034482776)(5,0.3001834433943965)(6,0.3099681771659486)(7,0.37857185882543076)(8,0.41663314760775866)(9,0.49871852795258664)};
\addplot+ [line4] coordinates
{(0,0.00390625)(1,0.2503854167504189)(2,0.04361700747068675)(3,0.297048605410385)(4,0.3026875382244558)(5,0.30150969926298177)(6,0.3197006980402011)(7,0.38352264834170874)(8,0.42121082579564484)(9,0.5125564644723618)};
\addplot+ [line5] coordinates
{(0,0.00390625)(1,0.2468540488771931)(2,0.04322231866666668)(3,0.30066200140350885)(4,0.3072547044210527)(5,0.30321313863157895)(6,0.32191579140350884)(7,0.38796111189473714)(8,0.42309090817543843)(9,0.5191748479649123)};
\addplot+ [sharp plot, mark=asterisk,mark size=1.5pt,mark options={mark color=brown}, color=brown] coordinates
{(0,0.00390625)(1,0.23370011909090904)(2,0.04343352808612441)(3,0.3003948985645933)(4,0.3030038201913876)(5,0.2943547575598084)(6,0.31507045229665076)(7,0.38787051928229666)(8,0.4143123010526318)(9,0.5222597723923448)};
\end{axis}
\end{tikzpicture}
\begin{tikzpicture}[
line1/.style={sharp plot,mark=o,mark size=1.5pt,mark options={mark color=red}, color=red},
line2/.style={sharp plot, mark=square*,mark size=1.5pt,mark options={mark color=blue}, color=blue},
line3/.style={sharp plot, mark=diamond*,mark size=1.5pt,mark options={mark color=cyan}, color=cyan},
line4/.style={sharp plot, mark=pentagon*,mark size=1.5pt,mark options={mark color=pink}, color=pink},
line5/.style={sharp plot, mark=triangle*,mark size=1.5pt,mark options={mark color=violet}, color=violet},
line7/.style={sharp plot, mark=asterisk,mark size=1.5pt,mark options={mark color=brown}, color=brown}
]
\tikzset{every node}
\begin{axis}
[width=5cm,enlargelimits=0.13, tick align=inside, 
xticklabels={ $2$,$ $,$4$,$ $,$6$,$ $,$8$,$ $,$10$,$ $,$ $},
xtick={0,1,2,3,4,5,6,7,8,9,10},
xlabel={Layers}]
\addplot+ [line1] coordinates
{(0,0.0078125)(1,0.14232394433333337)(2,0.03381585700000002)(3,0.1872079500833334)(4,0.23145063316666667)(5,0.2314355168333333)(6,0.19859951041666674)(7,0.2839086518333332)(8,0.3487794474999999)(9,0.37868058249999986)(10,0.3997015316666665)};\label{line1}
\addplot+ [line2] coordinates
{(0,0.0078125)(1,0.16655493292824075)(2,0.034511326747685166)(3,0.19185990407407397)(4,0.2624877119097223)(5,0.2515993033101851)(6,0.24664107185185175)(7,0.32525602547453686)(8,0.37769124690972194)(9,0.4115393387731481)(10,0.4283315676504632)};
\addplot+ [line3] coordinates
{(0,0.0078125)(1,0.17308842678879316)(2,0.03444576387931033)(3,0.19484727283405173)(4,0.26684479547413836)(5,0.2573370572306037)(6,0.2614769775431035)(7,0.34139276238146543)(8,0.38279373455818927)(9,0.4260450118857763)(10,0.44165199493534485)};
\addplot+ [line4] coordinates
{(0,0.0078125)(1,0.1754957419597989)(2,0.03431029604690116)(3,0.19647850616415416)(4,0.27157486105527634)(5,0.2617793521608039)(6,0.26727810361809046)(7,0.34659280296482425)(8,0.38514092805695144)(9,0.4355248534505859)(10,0.4538368710720269)};
\addplot+ [line5] coordinates
{(0,0.0078125)(1,0.17719729452631594)(2,0.03400829761403509)(3,0.19689670687719305)(4,0.273739137122807)(5,0.265638465649123)(6,0.27264950428070167)(7,0.3497031768070176)(8,0.3869416206315789)(9,0.44059771340350884)(10,0.45729987663157884)};
\addplot+ [sharp plot, mark=asterisk,mark size=1.5pt,mark options={mark color=brown}, color=brown] coordinates
{(0,0.0078125)(1,0.17632217531100475)(2,0.03406728746411483)(3,0.1959032183253589)(4,0.2672738021531099)(5,0.2647957341626795)(6,0.2665162794736842)(7,0.3436139494736842)(8,0.3724962059808613)(9,0.44424645349282277)(10,0.4558995787081342)};
\end{axis}
\end{tikzpicture}
\\
\begin{center}
   \ref{legend} 
\end{center}

\caption{Subgraph weights in models with different layers on WMT14 En-De. The first row shows models with 5, 6, 7 and 8 layers, and the second row shows models with 9, 10, 11, 12 layers.}
\label{fig:gateslength2}
\end{figure*}

\begin{figure*}[htp]
\centering
\subfigure[The first matrix for middle-order subgraphs.]{
\includegraphics[scale=0.33]{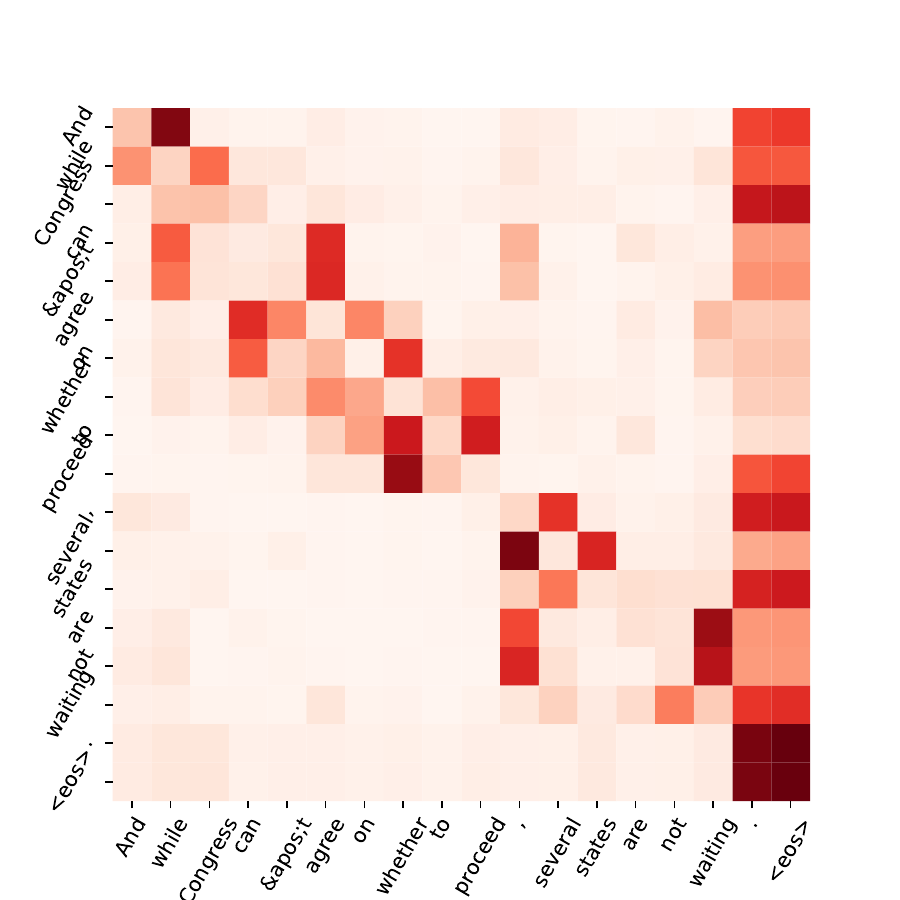}
}
\subfigure[The second matrix for middle-order subgraphs.]{
\includegraphics[scale=0.33]{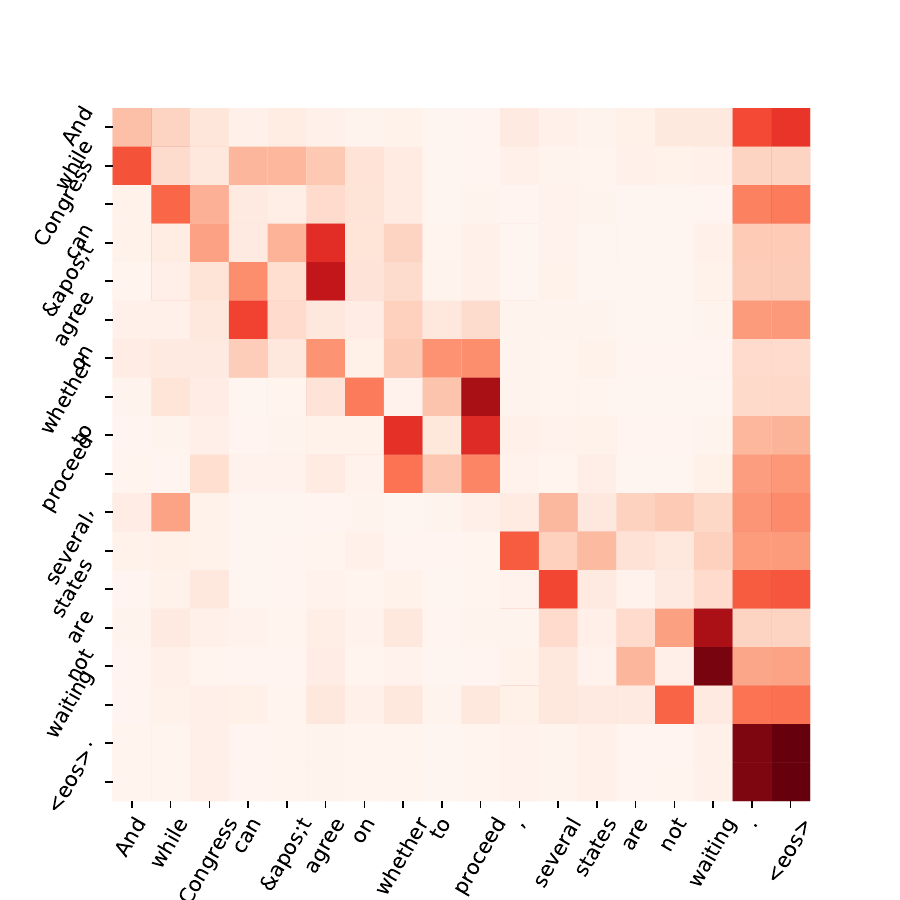}
}
\subfigure[The matrix for high-order subgraphs.]{
\includegraphics[scale=0.33]{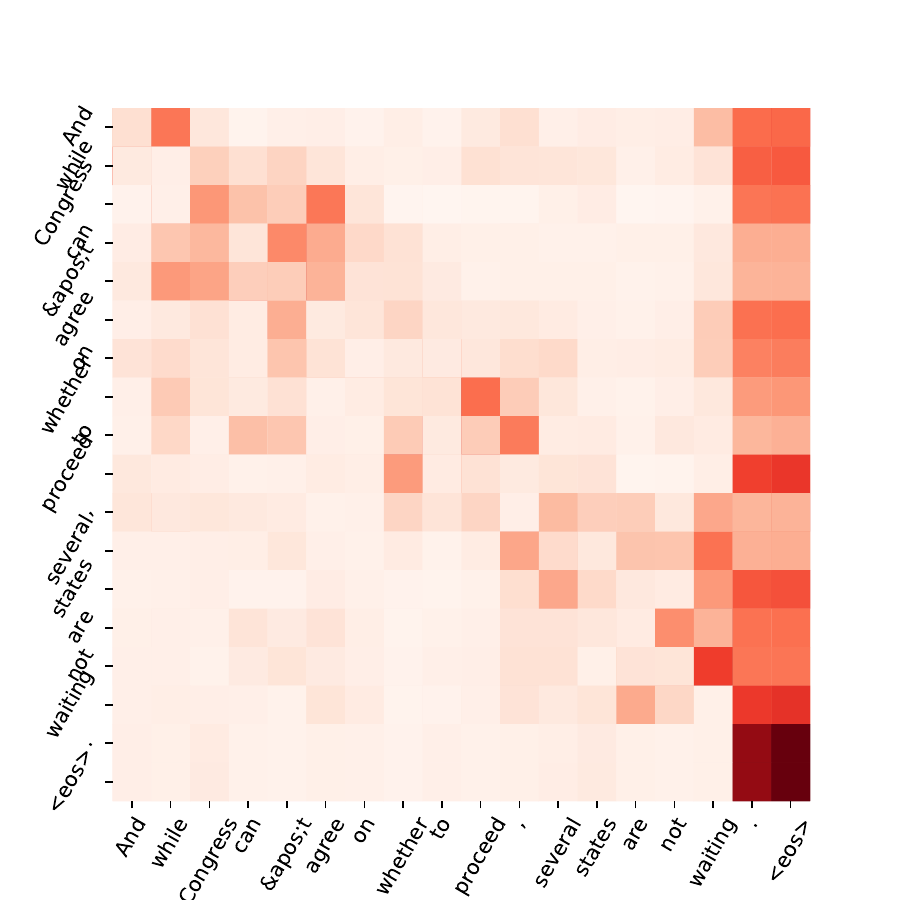}
}
\caption{Self-attention matrices of 6-th layer in WMT14 En-De model. }
\label{fig:ma}
\end{figure*}

Table \ref{bleu1ablation} shows the results of our Graph-Transformer in De-En and En-De tasks to evaluate our model with different methods. We evaluated three fusion strategies, i.e., \textbf{sum}, \textbf{weight-gate}, and \textbf{self-gate}. Table \ref{bleu1ablation} shows that \textbf{weight-gate} is the most effective among all fusing strategies. Using weight-gate to weight different groups of subgraphs has shown indeed helpful. We then evaluated the methods mentioned in Section \ref{section:41}, i.e., \textbf{half-dimension} (half-dim) and \textbf{shared-query-key-value} (shared-qkv), and found that the weight-gate+half-dim achieves the best performance on De-En and En-De.

Fig. \ref{fig:length} shows that our model outperform the baseline on all lengths. Especially, our model trends to be better than the baseline when the input sentences are longer than 20. It shows that our models performs better on longer sentences. Fig. \ref{fig:layer} shows that our model outperforms the baseline with different model depths. With the model depth increasing, the performance growth of our model is uniform.

To demonstrate the effect of sentence length and layer numbers on the weight of high-order subgraphs, we evaluate our models with different layer numbers using the half-dimension and weight-gate method on the En-De task and show the results in Fig. \ref{fig:gateslength2}. Fig. \ref{fig:gateslength2} reveals that longer sentences often require higher weights of high-order subgraphs than shorter sentences, no matter how many layers and at which layer of the model, and the weights usually increase at higher layers. Fig. \ref{fig:gateslength2} also shows that the weights drop at the fourth layer, meaning that subgraphs captured at the fourth layer are less important for the model than those captured in the previous three layers.

Fig. \ref{fig:ma} is the attention visualization for three parts of self-attentions and shows that the weight values in the matrix for subgraphs of high order are smoother than the values in the matrices for subgraphs of middle order. It means that the self-attention for subgraphs of high order focuses on relationships among more words and does capture subgraphs of high order.

\subsection{Analysis of Result}
According to the MoG explanation and the design of Graph-Transformer, not calculating subgraphs of low order can avoid generating subgraphs repeatedly. It ensures that every subgraph generates only once and with the same weight to improve model performance slightly.  Meanwhile, weighting subgraphs allows the model to figure out salient subgraphs. Without weighting subgraph, our model can only outperform the baseline by 0.4 BLEU points on the En-De task, and outperform the baseline by more than 0.9 BLEU points after weighting subgraphs using weight-gate. It is the same as we expected and indicates the reasonableness of our MoG explanation.

Table \ref{bleu1ablation} compares different fusion strategies, in which weight-gate performs best, while self-gate is not the best one. Calculating the sum of representation only makes every subgraph be generated once and have the same weight to stop the model from figuring out salient subgraphs, and performs worst. Self-gate can weigh every group of subgraphs which cannot be done by weight-gate. However, using self-attention and representation, self-gate will generate new subgraphs of high order and unnecessary redundancy. Self-gate also makes models deeper and difficult to train. Although weight-gate cannot distinguish every subgraph in representation, it makes the model focus on specific parts. When we produce representations using the same query, key, and value, there are some stable relationships between them. Dividing them into two groups can mostly distinguish this relationship and enable the model to capture it. Table \ref{bleu1ablation} also shows that the same method may perform differently for different tasks. For example, the shared-qkv method hurt the performance of weight-gate on De-En, while the weight-gate+half-dim+shared-qkv achieves a comparable performance compared with the weight-gate+half-dim on De-En.

Table \ref{bleu1ablation} shows that the model with half dimension can get a similar or better result than the model with full dimension. A larger model dimension enables the vector to accommodate more features. Our results do not mean a larger dimension is unimportant. Though we use fewer parameters, our model can capture subgraphs more accurately. Our model can distinguish subgraphs of different orders with three independent parts of self-attention. The half dimension  generated by a non-linear operation removes some subgraphs from the representations and pushes our model to focus on other subgraphs, which enhances the ability of our model to select subgraphs of higher order. Besides, more parameters usually make the model more difficult to train and easier to be overfitting. Thus the half dimension setting helps the resulted model to outperform one with full dimension.

\begin{table}[htb]
\caption{Results of text summarization.}
\small

    \centering
    \begin{tabular}{l|c|c|c}
    \hline
    
&ROUGE-1&	ROUGE-2&	ROUGE-L\\
\hline
Transformer	&36.84	&18.01&	34.31\\
\hline
Our Model&	37.60 (+0.76) &	18.63 (+0.62) &	34.71 (+0.4)\\
        \hline
    \end{tabular}
    
    \label{results2}
\end{table}

\subsection{Evaluation on Text summarization}

We evaluate our model with half-dimension+weight-gate on Text summarization (SUM) tasks using the same training set as WMT14 En-De. We use the Annotated Gigaword dataset with 3.8M sentence pairs for the SUM task training, and BPE algorithm to process words into subwords with 32K tokens. Table \ref{results2} shows that our model with half-dimension+weight-gate outperforms the baseline on all evaluation metrics.

\subsection{Evaluation on GLUE}

\begin{table*}[!ht]
\caption{Results of GLUE with fine-tuned \texttt{bert-base-cased}.}
    \centering
    \begin{tabular}{l|ccccccccc}
    \hline
    \multirow{2}*{Model} &CoLA&SST-2&STS-B&RTE&QNLI&MNLI&QQP&MRPC&WNLI\\
    &(mc)&(acc)&(pc/sc)&(acc)&(acc)&m/mm(acc)&(acc/F1)&(F1/acc)&(acc)\\
        \hline
BERT &50.4 &93.2 &85.2/83.4 &56.7 &90.3 &84.2/84.1 &89.1/71.6 &86.8/81.7&61.0 \\
\hline
Our Model & 54.3& 93.8 & 86.5/85.2& 61.9& 90.7& 84.6/83.9&89/71.6&88.8/84.4&65.8\\
\hline
    \end{tabular}
    
    \label{glue}
\end{table*}

\begin{table*}[!ht]
\caption{Results of GLUE with frozen \texttt{bert-base-cased}.}

    \centering
    \begin{tabular}{l|ccccccccc}
    \hline
    \multirow{2}*{Model}&CoLA&SST-2&STS-B&RTE&QNLI&MNLI&QQP&MRPC&WNLI\\
    &(mc)&(acc)&(pc/sc)&(acc)&(acc)&m/mm(acc)&(acc/F1)&(F1/acc)&(acc)\\

        \hline
BERT &39.2 &90.2 &79.9/77.6 &57.6 &86.0 &79.0/78.9 &87.3/68.2 &82.1/73.2&57.5 \\
\hline
Our Model & 41.7& 90.9 & 82.2/80.6& 59.2& 86.6& 80.5/79.8&87/68.2&84.4/77.4&57.5\\
\hline
    \end{tabular}
    
    \label{gluefrozen}
\end{table*}

To evaluate our model on GLUE \cite{wang2019glue} tasks, we first train a pre-trained language model and then fine-tune the pre-trained language model for GLUE tasks. For fairly compared to the baseline, we add six random initialized BERT layers on one \texttt{bert-base-cased} as baseline pre-trained language model and add sex random initial layers of Graph-Transformer with half-dimension+weight-gate  on another BERT as our pre-trained language model. The data for training is wikipedia \footnote{\url{https://huggingface.co/datasets/wikipedia}}, and the pre-processed subset is \textit{20220301.en}. We use the same model config of \texttt{bert-base-cased} for our model and the baseline. To train the pre-trained language model, the batch size is 128, the learning rate is 5e-5, the maximum length is 128. We train the pre-trained language model for 1 epoch. The implementation of our model and the baseline is based on transformers \cite{wolf-etal-2020-transformers}.  The script for training is from transformers \footnote{https://github.com/huggingface/transformers/blob/main/examples/pytorch\\/language-modeling/run\_mlm\_no\_trainer.py}. 

For GLUE tasks, we fine-tune our pre-trained language model for different tasks, and the script for training is from transformers \footnote{https://github.com/huggingface/transformers/blob/main/examples/pytorch\\/text-classification/run\_glue.py}. We train our model on different tasks for 3 epochs (5 epochs on WNLI and MRPC) with learning rate 2e-5. Table \ref{glue} shows the results of GLUE tasks, and shows that our model can improve the performance of CoLA, SST-2, STS-B, RTE, QNLI, MNLI, MRPC and WNLI, and get a comparable result on QQP which does not hurt the performance. 

We also train our pre-trained language model and the baseline pre-trained language model using frozen \texttt{bert-base-cased} without change of training and model config to avoid the effect of \texttt{bert-base-cased} and compare our model with the baseline more fairly. Table \ref{gluefrozen} shows the results of GLUE tasks with frozen \texttt{bert-base-cased}, and shows that our model can improve the performance of CoLA, SST-2, STS-B, RTE, QNLI, MNLI and MRPC, and get comparable results on QQP and WNLI.

\section{Conclusions}
This paper presents a unified explanation for representations given by SAN-based encoders, especially, the SAN empowered Transformer. Instead of a simple directed graph modeling in previous work, we re-define multigraph into Multi-order-Graph to accommodate broad categories of complicated relationships inside the representations. MoG connects not only words but also subgraphs. With the built relationship by MoG, we can understand diverse relationships inside representations in a unified way. Inspired by the proposed MoG explanation, we further propose a Graph-Transformer to enhance the ability to capture subgraph-aware representations on the SAN-based encoder. Experimental results indicate that our proposed MoG explanation for representations is empirically reasonable.

\bibliographystyle{IEEEtran}
\bibliography{pami}

%




\end{document}